\PassOptionsToPackage{table,dvipsnames}{xcolor}
\documentclass{article}
\PassOptionsToPackage{numbers, compress}{natbib}

\usepackage[preprint]{neurips_2025}

\usepackage[utf8]{inputenc} 
\usepackage[T1]{fontenc}    
\usepackage{hyperref}       
\usepackage{url}            
\usepackage{booktabs}       
\usepackage{amsfonts}       
\usepackage{nicefrac}       
\usepackage{microtype}      

\title{Are Large Language Models Reliable AI Scientists? \\ Assessing Reverse-Engineering of Black-Box Systems}

\author{%
  Jiayi Geng\thanks{Equal contribution}\\
  Department of Computer Science\\
  Princeton University\\
  \texttt{jiayig@princeton.edu}
  \And
  Howard Chen\footnotemark[1] \\
  Department of Computer Science \\
  Princeton University \\
  \texttt{howardchen@cs.princeton.edu}
  \AND
  Dilip Arumugam \\
  Department of Computer Science\\
  Princeton University\\
  \And
  Thomas L. Griffiths \\
  Department of Computer Science\\
  Department of Psychology \\
  Princeton University\\
}

\usepackage{amsmath}
\usepackage{bbm}
\usepackage{wrapfig}
\usepackage{graphicx}
\usepackage{multirow} 
\usepackage{makecell}
\usepackage{fvextra}            
\usepackage[table, dvipsnames]{xcolor}

\definecolor{instancepink}{RGB}{255, 220, 220}
\definecolor{modelgray}{RGB}{245,245,245}

\usepackage{tcolorbox}
\usepackage{listings}
\definecolor{instancepink}{RGB}{255, 220, 220}
\definecolor{modelbox}{RGB}{245,245,245}
\definecolor{myred}{RGB}{180,0,0}

\DefineVerbatimEnvironment{code}{Verbatim}{
  breaklines,                   
  breakanywhere=true,   
  breaksymbolleft={},  
  breaksymbolright={},
  breakindent=0pt,
}

\begin{document}

\maketitle
\begin{abstract}
Using AI to create autonomous researchers has the potential to accelerate scientific discovery. A prerequisite for this vision is understanding how well an AI model can identify the underlying structure of a black-box system from its behavior. 
In this paper, we explore how well a large language model (LLM) learns to identify a black-box function from passively observed versus actively collected data.
We investigate the reverse-engineering capabilities of LLMs across three distinct types of black-box systems, each chosen to represent different problem domains where future autonomous AI researchers may have considerable impact: programs, formal languages, and math equations. 
Through extensive experiments, we show that LLMs fail to extract information from observations, reaching a performance plateau that falls short of the ideal of Bayesian inference.
However, we demonstrate that prompting LLMs to not only observe but also intervene---actively querying the black-box with specific inputs to observe the resulting output---improves performance by allowing LLMs to test edge cases and refine their beliefs.
By providing the intervention data from one LLM to another, we show that this improvement is partly a result of engaging in the process of generating effective interventions, paralleling results in the literature on human learning.
Further analysis reveals that engaging in intervention can help LLMs escape from two common failure modes: \textit{overcomplication}, where the LLM falsely assumes prior knowledge about the black-box, and \textit{overlooking}, where the LLM fails to incorporate observations.
These insights provide practical guidance for helping LLMs more effectively reverse-engineer black-box systems, supporting their use in making new discoveries. Codes are available at \url{https://github.com/JiayiGeng/reverse-engineering}.
\end{abstract}

\section{Introduction}
\label{sec:introduction}
\begin{figure}[h]
    \centering
    \includegraphics[width=\columnwidth]{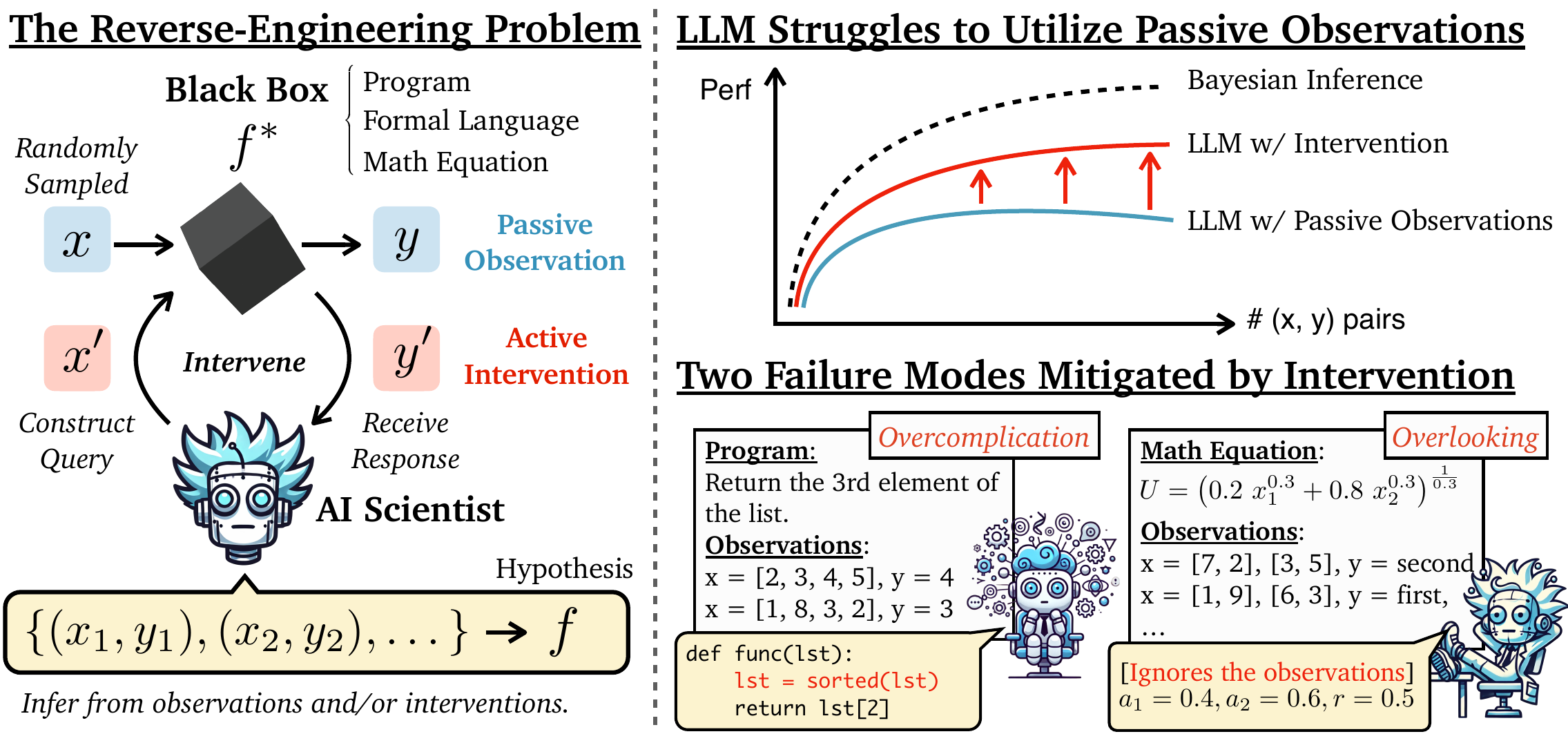}
    \caption{\textbf{Reverse-engineering.}
        Left: Defining the problem. The AI scientist will obtain either passive observations from the black box or collect data through active intervention to construct a hypothesis. Right (top): with only passive observations, the LLM cannot make effective use of the data and lags behind Bayesian inference by large margin; allowing the LLM to intervene improves performance. Right (bottom): effective intervention can mitigate two common failure modes: overcomplication and overlooking.
    }
    \label{fig:teaser}
\end{figure}
\vspace{-2pt}
Developing intelligent systems to accelerate scientific discovery has been a long-standing goal of artificial intelligence research~\citep{gil2014amplify,wang2023scientific}.
Despite rapid progress in creating large language models (LLMs) for understanding text and solving problems such as math and coding, automating scientific processes poses a different kind of challenge.
A core aspect of scientific discovery is \textit{reverse-engineering} the underlying mechanism behind a black-box system, which requires capabilities beyond responding to a one-off query.
In particular, reverse-engineering often involves 1) understanding a collection of observed data in order to develop hypotheses, 2) designing experiments to actively acquire informative data from the black-box to test those hypotheses, and 3) describing and communicating the results.

Existing work using LLMs for automating scientific processes either focuses on static observational data \citep{rmus2025towards, shojaee2025llm} or emulates scientific workflows using ``LLM scientists'' with many moving parts \citep{gandhi2025boxinggym, schmidgall2025agent}. 
In contrast, research in related fields has used carefully controlled tasks to evaluate whether machine learning systems can perform key aspects of reverse-engineering, including inductive reasoning \citep{rule2024symbolic}, learning causal features from passive data \citep{lampinen2023passive}, and optimal experimental design \citep{chaloner1995bayesian,foster2019variational}. 
This work is often informed by research in cognitive science, which has studied how humans engage in active learning using methods in which the source (\textit{i.e.} passive observation or active experimentation) and content of data can be differentiated \citep{markant2010category,markant2014better}.
However, such controlled methodologies have not yet been applied to evaluating state-of-the-art LLMs, leaving fundamental questions unanswered: ``\textit{How well can LLMs make inferences from passive observations?}'' and ``\textit{Can they actively collect data to refine their hypotheses?}''.

To answer these questions, we systematically study LLMs on three reverse-engineering tasks inspired by the cognitive-science literature and selected to mimic challenges that arise in scientific settings: reconstructing list-mapping programs \citep{rule2024symbolic}, formal languages \citep{mccoy2023modeling}, and math equations \citep{foster2019variational}.
Through extensive experiments, we show that LLMs are limited in their ability to make inferences from observations, leading to performance plateaus when compared to Bayesian models.
However, allowing LLMs to perform interventions---generating test cases or queries to collect new, informative data---can significantly improve their performance.

Through further experiments in which the results of the interventions conducted by one LLM become observational data for another, we show that the benefits of intervention seem to come from the LLM testing and refining its own beliefs rather than simply collecting higher-quality data. This is similar to a phenomenon observed in human learning, where people show limited benefit from interventions generated by others \citep{markant2010category, markant2014better}. Further investigation reveals that generating  interventions seems to help LLMs overcome two failure modes: 1) \textit{overcomplication}, where the LLM tends to construct overly-complex hypotheses, and 2) \textit{overlooking}, where the LLM neglects observations or draws overly-generic conclusions without careful checking.


Our contributions are as follows:
\begin{itemize}
    \item Drawing inspiration from controlled studies of human cognition, we formalize \textit{reverse-engineering} as a core problem for assessing the scientific discovery capabilities of LLMs and design three black-box tasks that can be used in such assessment.
    \item We demonstrate empirically that frontier LLMs still struggle, relative to Bayesian inference, at reverse-engineering these black boxes when provided with only passive observations.
    \item We show that LLMs can perform interventions to obtain more informative data, and that effective intervention mitigates the failure modes of \textit{overcomplication} and \textit{overlooking}.
    \item We show that performance degrades when repurposing the LLM's intervention data as observations, pinpointing the mechanism behind the improvements it produces and highlighting a potential pitfall for exchanging knowledge among LLMs.
\end{itemize}
\section{Related Work}
\label{sec:related_work}

\paragraph{Inductive Inference}
Some of the earliest work on reverse-engineering appears under the label of \emph{inductive inference} for ``hypothesizing a general rule from examples''~\citep{angluin1983inductive}. Classic instances of this problem include work on identifying the underlying structure of a finite-state automaton through observations of its input-output behavior~\citep{rivest1987diversity,rivest1989inference}. While this problem typically considers passive observations, seminal work on active learning focuses on analyzing the benefits of actively querying inputs to solicit the most-informative outputs from the unknown function of interest~\citep{littlestone1988learning,angluin1988queries,settlestr09}. The key distinction between these seminal works and ours is the attention towards LLMs and assessing their capacity for successfully identifying different types of black boxes from input-output examples. 

\paragraph{Bayesian Optimal Experiment Design}
An adjacent line of work considers the sequential design of experiments which maximally yield information gain about an unknown parameter of interest~\citep{lindley1956measure,degroot1962uncertainty,chaloner1995bayesian,foster2019variational}; one may interpret these methods as studying a non-LLM-focused, Bayesian analogue of the reverse-engineering problem we formulate in the subsequent section, where a learner begins with a prior distribution over the black box in question and must maximally reduce epistemic uncertainty~\citep{der2009aleatory} with a given budget of experiments. To the extent that LLMs may implicitly engage with an underlying approximate posterior inference scheme~\citep{xie2021explanation,griffiths2024bayes,zhu2024eliciting,falck2024context,mccoy2024embers}, the reverse-engineering capabilities studied in this work can be tied to this Bayesian optimal experiment design problem.

\paragraph{Reinforcement Learning}
The fundamentals of the reverse-engineering problem also connect with various ideas studied in the context of reinforcement learning (RL)~\citep{sutton1998introduction}. Any model-based RL agent~\citep{sutton1990integrated,sutton1991dyna,brafman2002r,strehl2008analysis} naturally engages with a particular instance of the reverse-engineering problem where the black-box function in question is the transition function and/or reward function of a Markov Decision Process (MDP)~\citep{bellman1957markovian,Puterman94}.
The distinction explored in this work between a LLM that passively observes versus actively intervenes on the black box in question has a direct connection to the exploration challenge in RL, which has profound impact on an agent's ability to recover an accurate model of the world~\citep{thrun1991active,deisenroth2011pilco,strens2000bayesian,osband2013more}; while recent work~\citep{arumugam2025toward} has studied how to improve exploration with LLMs, this paper focuses on assessing the innate capabilities of LLMs to actively query informative data.  The KWIK learning framework of \citet{li2008knows} provides a theoretical analysis for reverse-engineering a MDP transition function when a learner must either confidently estimate the environment dynamics or say ``I don't know''~\citep{walsh2009exploring,li2010reducing,sayedi2010trading,szita2011agnostic,abernethy2013large}. 
Finally, there is a connection between intervention for effective reverse-engineering and meta RL~\citep{liu2021decoupling}, with recent work showing that passive learning can be effective with LLMs once there is an effective exploration strategy capable of yielding high-quality observations~\citep{lampinen2023passive}; naturally, the latter problem is precisely what we demonstrate interventions allow LLMs to solve for themselves in reverse-engineering tasks.

\paragraph{LLMs for Automating the Scientific Process} 
With the rapid advances in LLMs, recent work has explored using them to automate different parts of the scientific process such as ideation \citep{si2024can}, assistance \citep{gottweis2025towards}, writing research papers \citep{lu2024the, starace2025paperbench}, or emulating AI scientists in simulated environments \citep{schmidgall2025agent}. Additionally, multi-modal and multi-agent AI models have driven significant progress in applications such as protein science \citep{o2025sparks}, while frameworks like MatPolit \citep{ni2024matpilot} integrate human cognitive insights to accelerate discoveries in materials science. These works utilize the abundant knowledge stored in the LLMs to directly tackle real-world complexity in science \citep{reddy2025towards}. However, the complexity of these settings and the resulting agents make it hard to disentangle the consequences of all the engineering choices that go into these systems. Our work instead focuses on using simple and controllable black boxes to study the core capabilities of the LLMs themselves.

\paragraph{Understanding Failure Modes in LLMs}
Recently, many works have examined the failure modes of formal reasoning in LLMs. It has been observed that LLMs can exhibit failure modes of both ``overthinking'' \citep{chen2024not} and ``underthinking'' \citep{wang2025thoughts} when tackling mathematical problems and code generation \citep{he2025can, sprague2024cot, cuadron2025danger, sprague2024cot, sui2025stop, cemri2025multi}.
To understand LLM abilities beyond formal reasoning tasks, recent work has leveraged insights and datasets from cognitive science \citep{frank2023baby,binz2023using,coda2024cogbench,ying2025benchmarking}. In particular, researchers have started to use cognitive science to explore the failed behaviors in LLMs \citep{ku2025using}. Using these methods, researchers have found that LLMs sometimes overestimate human rationality \citep{liu2024large},  exhibit inconsistencies in probability judgments \citep{zhu2024incoherent}, and perform worse as a result of engaging in reasoning  \citep{liu2024mind}. In a similar vein, our work draws upon research from cognitive science to design the black boxes used in our reverse-engineering experiments. 

\section{Reverse Engineering}
\label{sec:reverse_engineering}

\subsection{Problem Formulation}\label{sec:problem_formulation}
We define a black box $f^*: \mathcal{X} \to \mathcal{Y}$ as a deterministic function that maps a query $x \in \mathcal{X}$ to a response $y \in \mathcal{Y}$ through its internal dynamics. The \textbf{reverse-engineering} problem is for a model to infer the internals of a black box $f^*$ (\textit{e.g.} list mapping programs, production rules of formal languages, and math equations) from a sequence of query-response pairs $\mathcal{O} = \{ (x_1, y_1), (x_2, y_2), \ldots, (x_N, y_N) \} \subset \mathcal{X} \times \mathcal{Y}$ (Figure~\ref{fig:teaser}).
We consider two cases of the reverse-engineering problem: \textbf{observation-only} and \textbf{observation-intervention}.
In the observation-only scenario, all the queries are \textit{randomly sampled} from $\mathcal{X}$ and the corresponding response $y_i = f^*(x_i)$ is generated by the black box from a uniform distribution to construct the observation set. A large language model $\mathcal{M}$ must generate a hypothesis $f = \mathcal{M}(\mathcal{O})$ without further interaction with the black box. This setting assesses the model's ability to perform inductive reasoning \citep{angluin1983inductive}. 
In the observation-intervention scenario, the LLM is first given a set of observations $\mathcal{O}$ obtained in the observation-only scenario and is instructed to interact with the black box in a multi-round fashion. In each round, the LLM chooses one of the following actions: 1) construct a new query $x_{N+1}$ to query the black box and obtain the response $y_{N+1}$, 2) construct a new query-response pair $(x_{N+1}, y'_{N+1})$ and check its validity using the black box ($\mathbbm{1}[ y'_{N+1} = f^*(x_{N+1}) ]$), or 3) stop and conclude with a hypothesis $f$ about the black box. Before constructing the new query, the LLM can analyze the current oservations with strategies such as verbalizing its current belief or describing the current hypothesis (\S\ref{sec:intv_strategy}).
Before the LLM chooses to stop or reaches the maximal number of rounds, the query-response pairs obtained during intervention are appended to $\mathcal{O}$ for the next round.

\subsection{Black-Box Types}\label{sec:black_box_types}

Drawing on the literature on inductive inference in cognitive science, we select tasks commonly used to study learning of complex relationships to design our black-box systems and scale them up for evaluation with LLMs. These three distinct black-box function classes -- Program, Formal Language, and Mathematical Equation -- simulate problems encountered in scientific reverse-engineering scenarios. Due to space constraints, detailed black-box designs are relegated to Appendix \ref{sec:blackbox_design}.



\paragraph{Program.} We use list-mapping programs  \citep{rule2024symbolic} for the Program black-box. Each program implements a lambda expression (e.g., \texttt{(lambda(singleton(third \$0)))}) in Python, where the query is a list of integers and the response is an integer.

\paragraph{Formal Language.} The Formal Language black-box is defined by a simple program that generates sequences of symbols. For example, the language $A^n B^n$ generates sequences consisting of some number of $A$s followed by the same number of $B$s. The black-box allows the LLM to intervene by validating if a string is allowable under the rule. We define $46$ distinct black boxes each based on a language from \citet{yang2022} or \citet{mccoy2023modeling}.

\paragraph{Math Equation.} We use the Constant Elasticity of Substitution (CES) formulation from economics \citep{foster2019variational} as the Math Equation black-box. The utility $U = \left(\sum_i a_i x_i^r\right)^{\frac{1}{r}}$ is given by the weights $a_i$, the ratio $r$, and the quantities of each kind of goods $x_i$. The LLM can query the black-box with two lists of item types with quantities and obtain a response indicating which list has higher utility. 


\subsection{Evaluation Protocol}\label{sec:eval_protocol}
A black-box can be represented in multiple ways, rendering evaluation challenging. For example, two black-boxes can be compared through their descriptions in natural language (descriptive evaluation) or whether they respond similarly to the same queries (functional evaluation; see \S\ref{sec:func_eval}).
In this paper we focus on \textbf{descriptive evaluation}, where the black-box $f^*_{\text{NL}}$ is expressed in natural language, due to its communicative nature and  real-world use \citep{chopra2019first, gandhi2025boxinggym}. The LLM-generated hypothesis $f_\text{NL}$ is scored by an LLM judge against the black-box on a $0-10$ scale based on the criteria of each black-box type ($\texttt{score} = \texttt{LM-Judge}(f_{\text{NL}}, f^*_{\text{NL}})$). We use descriptive evaluation for Program and Formal Language. As the Math Equation does not require verbalization beyond the weights and ratio, we report the flipped root mean square error (1 - RMSE) between the inferred parameters and ground truth. 

\section{Experiments}
\label{sec:experiments}

\paragraph{Experimental setup.}
We use different versions of GPT-4o \citep{hurst2024gpt} for reverse-engineering (\textit{gpt-4o-2024-08-06}, dubbed as reverse-engineer LLM) and as the judge (\textit{gpt-4o-2024-05-13}, dubbed as the judge LLM). We use greedy decoding of both the reverse-engineer and the judge LLMs and report performance over $3$ seeds.
For the observation-only experiments, we report performance for number of observations $N = \{2, 5, 10, 20, 50, 100\}$.
For the observation-intervention setting, the reverse-engineer LLM performs $M = \{5, 10, 20, 50\}$ rounds of interventions conditioned on the initial set of $10$ observations ($\lvert \mathcal{O} \rvert = 10$).
In addition to GPT-4o, we report full results for Claude-3.5-Sonnet-20241022 \citep{anthropic2024claude35sonnet}, DeepSeek-R1 \citep{guo2025deepseek}, Llama-3.3-70B-Instruct \cite{grattafiori2024llama3herd} in Appendix \ref{app:diff_llm}.
We provide the prompts for both intervention and hypothesis generation in Appendix \ref{sec:prompts}. We also study different evaluation approaches in Appendix \ref{sec:func_eval}.

\subsection{LLM Struggles to Utilize Observations Optimally}
\label{sec:observation}


\begin{figure}[h!]
    \centering
    \includegraphics[width=\textwidth]{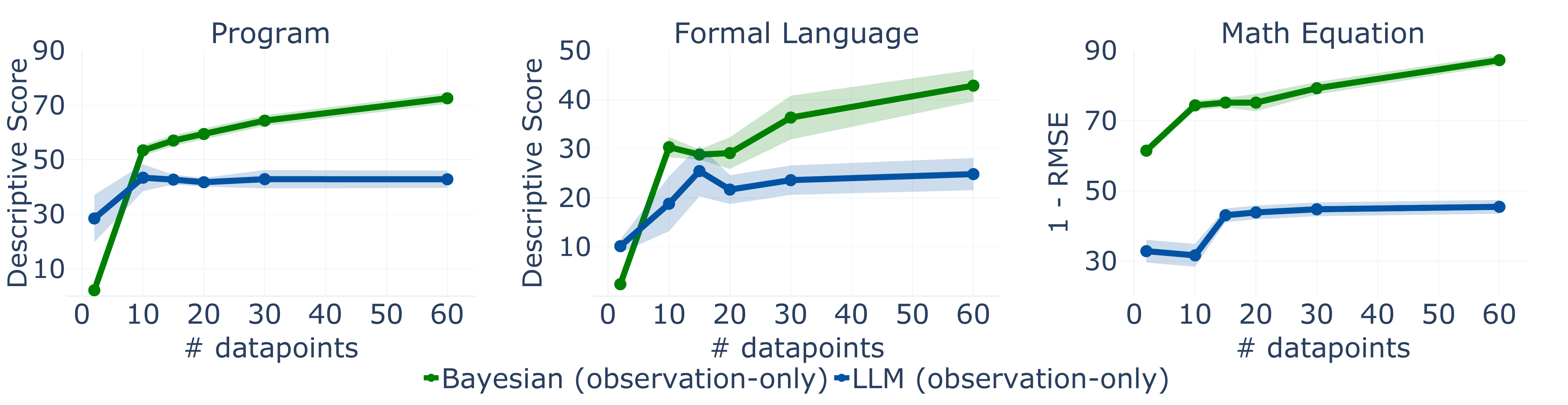}
    \caption{Observation-only results across three black-box types. We compare the GPT-4o performance (blue) to Bayesian inference (green).
    The horizontal-axis represents the number of provided $(x, y)$ pairs. We report 1 - RMSE for Math Equation and descriptive score for Program and Formal Language.
    }
    \label{fig:bayes_obs}
\end{figure}

We first establish the reference performance achievable by the Bayesian model in each setting. These three settings were selected in part because they are all cases where previous work has defined inference algorithms that make it possible to approximate the posterior distribution over hypotheses as more observations becomes available \citep{rule2024symbolic, yang2022, foster2019variational}.
As shown in Figure~\ref{fig:bayes_obs}, the Bayesian models (green) consistently improve with the increased number of observations across all three tasks.
On the other hand, while the LLM reverse-engineer (blue) starts off with higher performance for Program and Formal Language, potentially leveraging its prior knowledge, it peaks at $10$ observations and struggles to use the extra observations thereby causing performance to plateau.
We also calculate repeated measures ANOVAs \citep{girden1992anova} for each black-box type and found significant Model $\times$ number of datapoints interactions for Program ($F(5,10)=51.9$, $p<0.001$), Formal Language ($F(5,10)=11.8$, $p=0.001$), and Math Equation ($F(5,10)=8.7$, $p=0.002$), showing that the Bayesian inference algorithms increasingly outperformed LLMs with additional observations.
Details for the ANOVA calculations are provided in  Appendix \ref{sec:anova}.

\subsection{Intervention Is Crucial for the LLM to Refine Hypotheses}
\label{sec:intervention}

\begin{figure}[h]
    \centering
    \includegraphics[width=\textwidth]{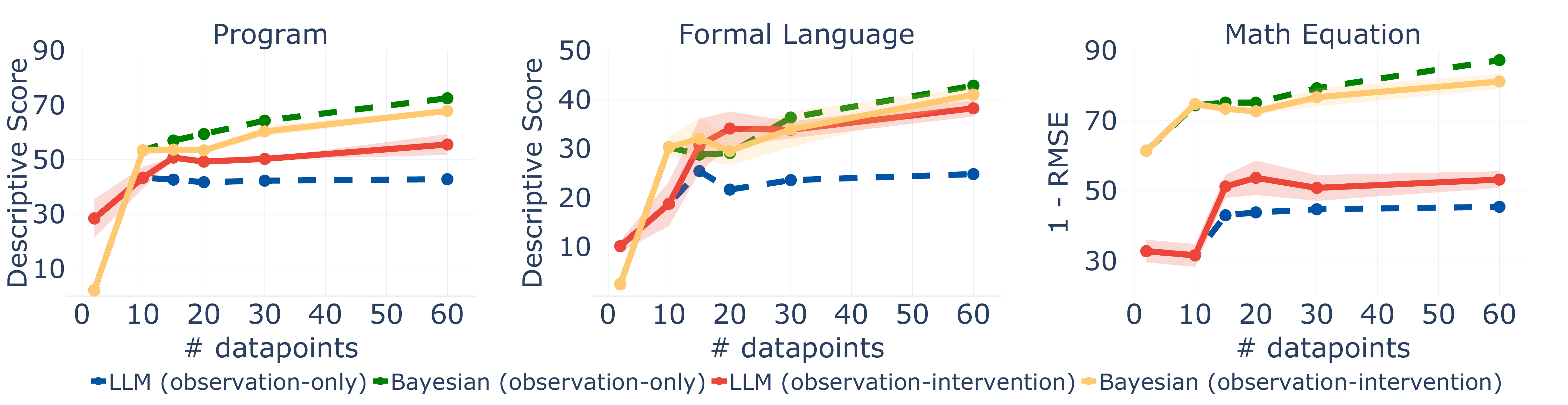}
    \caption{Observation-intervention results across three black-box types.
    Red: observations and interventions by GPT-4o. Yellow: taking the observation-intervention collected from GPT-4o as observations for the Bayesian inference algorithms.
    Dashed lines: observation-only reference for GPT-4o (blue) and Bayesian inference (green).
    }
    \label{fig:bayes_intv}
\end{figure}

In Figure~\ref{fig:bayes_intv}, we compare the performance of models with access to only the observations (dashed lines) against using the data that is actively collected through intervention (solid lines).
We observe that enabling the LLM to actively intervene significantly improves performance (red) over observation-only (dashed blue). Through intervention, the LLM consistently improves as more data becomes available across all three black-box types.
To assess the quality of the interventions, we provide the LLM-collected intervention data to the Bayesian model as observations, akin to the passive yoked data studied in \citet{markant2010category,markant2014better}. 
Our results indicate that while the interventions are beneficial to the LLM itself, they are not universally more informative, paralleling the findings in human active learning \citep{markant2010category,markant2014better}.
This gap was statistically significant, as shown by an ANOVA for each black box type: Program ($F(5,10)=23.9$, $p<0.001$), Formal Language ($F(5,10)=7.9$, $p=0.003$), and Math Equation ($F(5,10)=14.9$, $p<0.001$). 

\subsection{Identifying the Value of Generating the Intervention Data}
\begin{figure}[h!]
    \centering
    \includegraphics[width=\textwidth]{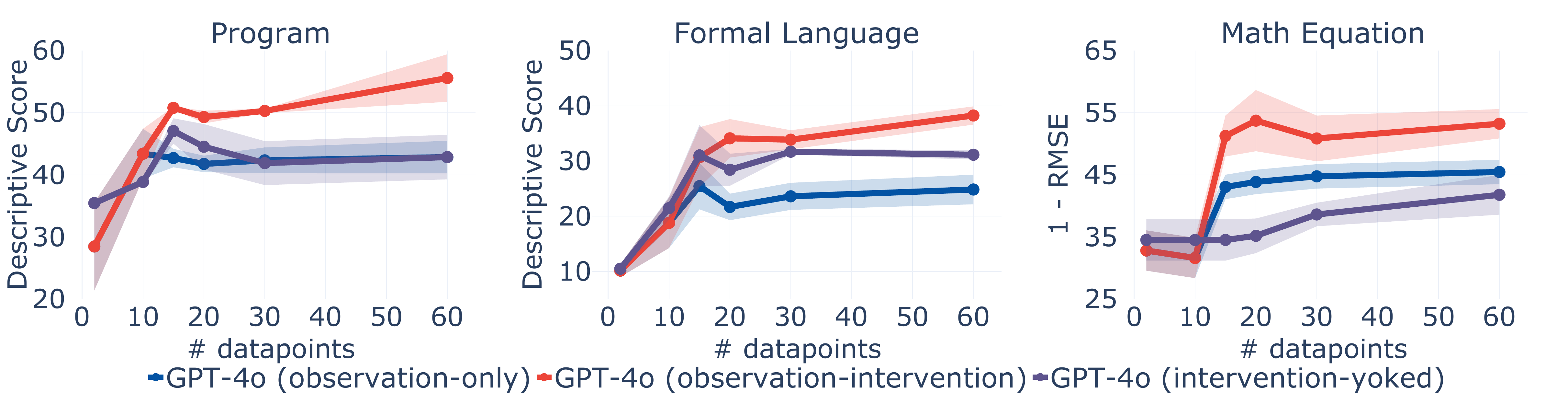}
    \caption{Comparing intervention-yoked results with observation-only and observation-intervention across three black-box types.}
    \label{fig:transfer_gpt4o}
\end{figure}

The improvement in performance produced by the interventions could have two sources: it could be that the resulting data are more informative, or that the process of generating interventions itself helps the model. To tease these apart, we adopt the passive-yoked design that \cite{markant2010category, markant2014better} used to study human learning, where the data generated via active learning by one group of participants is presented to another group of participants as passive observations.
In Figure~\ref{fig:transfer_gpt4o}, we compare GPT-4o across three conditions: \textbf{observation-only} (blue), \textbf{observation-intervention} (red), and \textbf{intervention-yoked} (purple) where GPT-4o only passively observes the interventional data without the verbalization and analysis that are used to construct such data. 
Results consistently show that the intervention-yoked setting leads to lower performance compared to the observation-intervention setting across all three black-box types. This shows that active learning is more beneficial than passive-yoked learning in part because it allows the LLM to dynamically refine its hypothesis in response to its own interventions.

\section{Analysis}
\label{sec:analysis}

\subsection{Escaping the Failure Modes: Overcomplication \& Overlooking}
\label{sec:failure_modes}
\begin{figure}[h!]
    \centering
    \includegraphics[width=\textwidth]{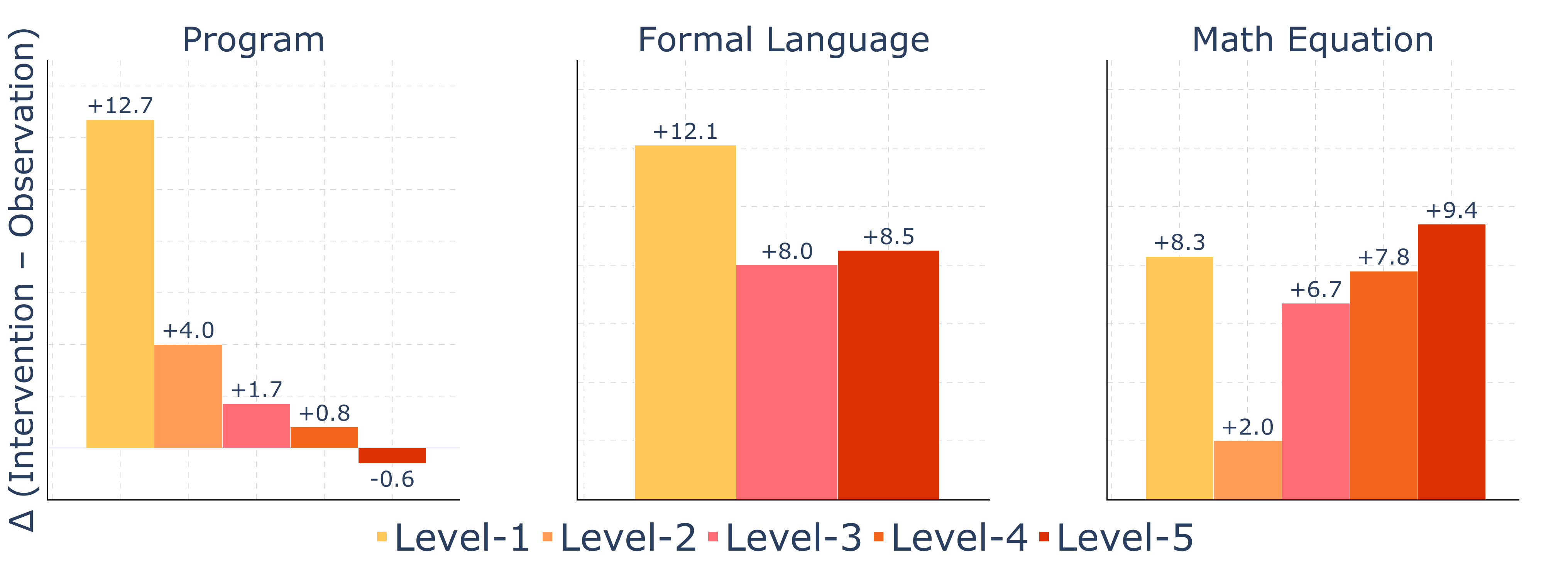}
    \caption{Descriptive scores for five different complexity levels. Averaged across three seeds for each of the three black-box types.}
    \label{fig:bar_plot}
\end{figure}

To understand how intervention improves LLM performance, we analyze common failures by sampling $20$ failed examples (scoring below $2$ out of $10$ points) from the observation-only experiment, which were inspected by human experts. We provide more details in Appendix \ref{sec:human_expert}. We identify two major failure modes: 1) \textit{overcomplication}, where the LLM excessively interprets the data, resulting in unnecessarily complex hypotheses, and 2) \textit{overlooking}, where the LLM inadequately leverages available information, leading to poorly reasoned hypotheses. 
We classified $20$ randomly sampled examples for each black-box into the two failure modes or ``Not Applicable'' by human annotation.
Results show that for Program the failures are  predominantly from overcomplication ($17$ cases out of $20$) whereas Math Equation contains more overlooking failures ($16$ cases out of $20$).
The failures are more evenly distributed for Formal Language, with $8$ examples classified as overcomplication, $11$ examples as overlooking, and $1$ example as ``Not Applicable''.
We provide examples for these failure mode in Appendix \ref{sec:overcomplication_overlooking_examples}.
Notably, we find that the impact of interventions on alleviating these two failure modes is contingent upon the complexity of the reverse-engineering task itself.
For each of the three specific domains we study, we include a brief characterization of complexity in Appendix \ref{app:complexity_cat}. Within each domain, we observe that the complexity of the reverse-engineering problem instance characterized by $f^*$ governs the extent to which interventions rectify failures of overcomplication and overlooking.
In Figure~\ref{fig:bar_plot}, we show that performance improvements from intervention on Program diminish as task complexity increases for black-box systems dominated by the \textit{overcomplication} failure mode. In contrast, actively collected data proves more beneficial when addressing challenging black-box instances dominated by the \textit{overlooking} failure mode, such as Math Equation. For Formal Language, where both failure modes frequently occur, we observe consistent improvements across all complexity levels.

\begin{figure}[h!]
    \centering
    \includegraphics[width=0.9\columnwidth]{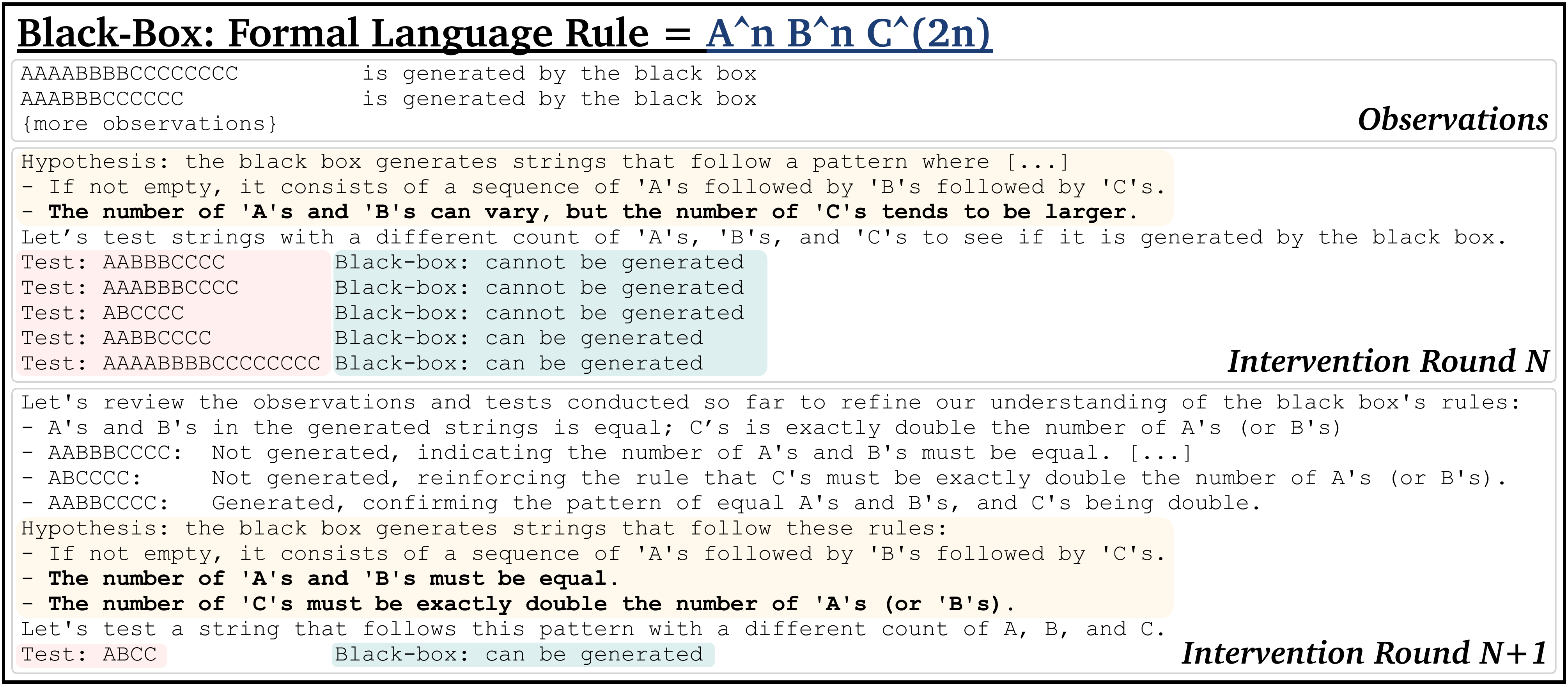}
    \caption{Case study example. GPT-4o updates the hypothesis using intervention on Formal Language black box.
    Yellow: GPT-4o verbalizes the hypothesis based on the passive observations in round $N$ and updates the hypothesis in round $N+1$. Red: constructing test cases. Teal: black box response.}
    \label{fig:intv_example}
\end{figure}

\paragraph{Case study.} 
Figure~\ref{fig:intv_example} demonstrates how an LLM progressively updates its hypothesis through active interventions to ultimately reverse-engineer the underlying mechanism of a black-box system using a Formal Language black-box intervention example, where GPT-4o strategically designs subsequent queries to validate its current belief about the system. In contrast, under the observation-only scenario, the model remains trapped in identifying spurious patterns from passively observed data and lacks a meaningful way to assess its own uncertainty. Through active interventions, the LLM iteratively tests and revises its hypotheses after encountering failures, gradually reducing uncertainty and converging toward an accurate understanding of the black-box mechanism.

\subsection{Intervention Strategies}\label{sec:intv_strategy}

Similar to how LLMs use chain-of-thought reasoning \citep{wei2022chain} to solve complex tasks, we allow the LLM to verbalize its hypotheses and analyze the observations before constructing the query. We investigate 

\begin{wraptable}{r}{0.67\linewidth}
  \centering
  \resizebox{0.99\linewidth}{!}{%
   \begin{tabular}{lcccc}
\toprule
&  & Descriptive & Functional & Analyze-then- \\
Black Box & Intervention & Intervention & Intervention & Query Intervention \\
\midrule
 Program & 43.4 & 47.6 & 19.2 & \textbf{50.8} \\
\midrule
 Formal Language & 24.1 & 28.6 & 22.8 & \textbf{34.7} \\
\midrule
 Math Equation & 34.8 & 38.8 & \textbf{39.9} & 38.0 \\
\bottomrule
\end{tabular}
}
\caption{Comparison of the four intervention strategies.}
\label{table:hypo_only}
\end{wraptable}
 how different reasoning  strategies impact the effectiveness of intervention. We compare four strategies: 1) Intervention: no reasoning before constructing the query, 2) Descriptive Intervention: describing the current hypothesis about the black-box, 3) Functional Intervention: verbalize the black-box implementation as a Python program \citep{li2025structured,luo2025deconstructing}, and 4) Analyze-then-Query: 
allowing the LLM to analyze data and verbalize a hypothesis freely. Throughout our experiments, we allow the LLM to reason once every five queries.

As shown in Table~\ref{table:hypo_only}, allowing the LLM to reason generally improves the effectiveness of intervention regardless of the strategy. However, the results also suggest that the most effective intervention typically requires the LLM to carefully analyze past observations and strategically plan subsequent steps to acquire more informative data from the black-box.
Interestingly, while structured reasoning in functional intervention \citep{li2025structured,luo2025deconstructing} is known to improve performance in formal reasoning tasks, it does not produce additional improvement in the context of reverse-engineering. This suggests that the LLM reverse-engineering abilities may differ from its formal reasoning capabilities.

\subsection{Transferring to another LLM}
\begin{figure}[h]
    \centering
    \includegraphics[width=\columnwidth]{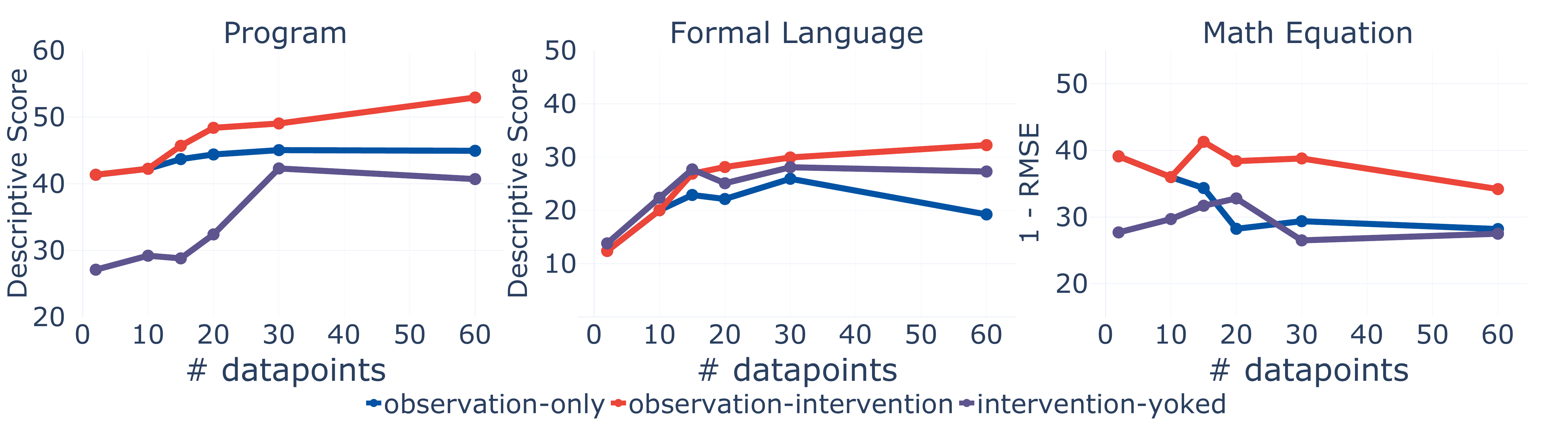}
    \caption{Intervention data transfer results. Red: Llama-3.3-70B-Instruct performing intervention. Blue: Llama-3.3-70B-Instruct using observations only. Purple: using interventional data from GPT-4o as observations for Llama-3.3-70B-Instruct.}
    \label{fig:transfer_llama}
\end{figure}
We also examine whether interventional data actively collected by one LLM (GPT-4o) can effectively transfer and benefit another LLM (Llama-3.3-70B-Instruct). This is relevant to whether AI scientists can transfer their experiments and findings successfully to another AI scientist.
Adopting a similar passive-yoked design, we compare three scenarios for Llama-3.3-70B-Instruct \citep{grattafiori2024llama3herd}: \textbf{observation-only}, \textbf{observation-intervention}, and \textbf{intervention-transfer}, where the interventional data is collected by GPT-4o.
As shown in Figure~\ref{fig:transfer_llama}, the intervention-transfer scenario achieves performance comparable to or slightly better than the observation-only baseline but consistently underperforms Llama's own intervention (observation-intervention).
This suggests that while the intervention data from GPT-4o is informative, the effectiveness diminishes when transferred to a different LLM, showing that the benefit from intervention is model-specific.

\section{Limitations and Future Directions}
We have discussed in this paper the LLM's inabilities and failure modes in reverse-engineering black-boxes. However, the three black-box types we studied represent only a narrow slice of possible tasks, even within controlled settings. A more comprehensive assessment will require expanding and scaling up the evaluation suite to probe LLMs’ reverse-engineering abilities across a broader spectrum of scenarios. In addition, we have assumed idealized, noise-free black-boxes and fully trustworthy data---a condition that is rarely met in real scientific practices. An important next step is to relax this assumption and rigorously test LLM robustness in the presence of noise and uncertainty.
As our paper discuss extensively on the failure modes of LLMs, we leave open the question: ``\textit{How can we train LLMs to become effective reverse engineers?}'', which includes enhancing the LLM's ability to perform correct inference from passive observations and to conduct optimal experiments.
In particular, what kinds of data and algorithms are needed to train such a model (e.g., reinforcement learning using black-box environments), and can improvements in one domain generalize to broader scientific automation tasks? 
Finally, we have demonstrated that the actively acquired data by one LLM may not be useful for another LLM, pointing to the issue of \textit{transferability of experiences}. This is important for automating scientific discovery as many major scientific advances have relied on effective collaborations. 
Understanding and quantifying the impact of this limited transferability of knowledge may be crucial as multi-agent systems become prevalent, and it will be essential to design such systems with effective communication baked in.
\section{Conclusion}
\label{sec:conclusion}
In this paper, we identified and formalized the reverse-engineering problem as a core ability and prerequisite for building a reliable AI scientists. We showed that current LLMs still struggle to effectively leverage passive observations even on seemingly simple and controlled black-boxes. Allowing LLMs to actively collect intervention data improves performance, but still falls short of closing the gap with Bayesian inference, casting doubt on the promise of truly reliable AI scientists. Through extensive analysis, we identified issues such as overcomplication and overlooking and illustrate how intervention can mitigate such failures. Despite the effectiveness of intervention, our analysis revealed that the intervention data collected by LLMs were primarily beneficial to the models themselves, rather than being objectively informative or transferable to other models. Altogether, our paper directly assesses the ability of LLMs to infer underlying causal structures and mechanisms through controlled reverse-engineering experiments. This capacity mirrors the fundamental scientific discovery process, which relies heavily on identifying hidden relationships and principles behind observed phenomena. Consequently, if an LLM cannot reliably reverse-engineer even simple or controlled systems, this raises concerns regarding its dependability in addressing more complex and ambiguous scientific challenges.
Evaluating an LLM’s reverse-engineering ability provides a concrete and principled way to assess its capacity for scientific reasoning, helping us understand whether such models possess the foundational skills required to function as dependable AI scientists.

\section*{Acknowledgments}
Our experiments were supported by Azure credits from a Microsoft AFMR grant. This work was supported by ONR MURI N00014-24-1-2748. Special thanks to Danqi Chen, Ryan Liu, Alexander Wettig, and Lucy He for providing invaluable feedback for this project.
\clearpage
\bibliography{neurips_2025}

\begin{thebibliography}{83}
\providecommand{\natexlab}[1]{#1}
\providecommand{\url}[1]{\texttt{#1}}
\expandafter\ifx\csname urlstyle\endcsname\relax
  \providecommand{\doi}[1]{doi: #1}\else
  \providecommand{\doi}{doi: \begingroup \urlstyle{rm}\Url}\fi

\bibitem[Abernethy et~al.(2013)Abernethy, Amin, Kearns, and Draief]{abernethy2013large}
Jacob Abernethy, Kareem Amin, Michael Kearns, and Moez Draief.
\newblock {Large-Scale Bandit Problems and KWIK Learning}.
\newblock In \emph{International Conference on Machine Learning}, pp.\  588--596, 2013.

\bibitem[Angluin(1988)]{angluin1988queries}
Dana Angluin.
\newblock {Queries and Concept Learning}.
\newblock \emph{Machine Learning}, 2:\penalty0 319--342, 1988.

\bibitem[Angluin \& Smith(1983)Angluin and Smith]{angluin1983inductive}
Dana Angluin and Carl~H Smith.
\newblock {Inductive Inference: Theory and Methods}.
\newblock \emph{ACM Computing Surveys (CSUR)}, 15\penalty0 (3):\penalty0 237--269, 1983.

\bibitem[Anthropic(2024)]{anthropic2024claude35sonnet}
Anthropic.
\newblock Claude 3.5 sonnet.
\newblock \url{https://www.anthropic.com/news/claude-3-5-sonnet}, 2024.

\bibitem[Arumugam \& Griffiths(2025)Arumugam and Griffiths]{arumugam2025toward}
Dilip Arumugam and Thomas~L Griffiths.
\newblock {Toward Efficient Exploration by Large Language Model Agents}.
\newblock \emph{arXiv preprint arXiv:2504.20997}, 2025.

\bibitem[Bellman(1957)]{bellman1957markovian}
Richard Bellman.
\newblock {A Markovian Decision Process}.
\newblock \emph{Journal of Mathematics and Mechanics}, pp.\  679--684, 1957.

\bibitem[Binz \& Schulz(2023)Binz and Schulz]{binz2023using}
Marcel Binz and Eric Schulz.
\newblock Using cognitive psychology to understand {GPT}-3.
\newblock \emph{Proceedings of the National Academy of Sciences}, 120\penalty0 (6):\penalty0 e2218523120, 2023.

\bibitem[Blei et~al.(2017)Blei, Kucukelbir, and McAuliffe]{blei2017variational}
David~M Blei, Alp Kucukelbir, and Jon~D McAuliffe.
\newblock Variational inference: A review for statisticians.
\newblock \emph{Journal of the American statistical Association}, 112\penalty0 (518):\penalty0 859--877, 2017.

\bibitem[Brafman \& Tennenholtz(2002)Brafman and Tennenholtz]{brafman2002r}
Ronen~I Brafman and Moshe Tennenholtz.
\newblock {R-MAX -- A General Polynomial Time Algorithm for Near-Optimal Reinforcement Learning}.
\newblock \emph{Journal of Machine Learning Research}, 3:\penalty0 213--231, 2002.

\bibitem[Brown et~al.(2020)Brown, Mann, Ryder, Subbiah, Kaplan, Dhariwal, Neelakantan, Shyam, Sastry, Askell, Agarwal, Herbert-Voss, Krueger, Henighan, Child, Ramesh, Ziegler, Wu, Winter, Hesse, Chen, Sigler, Litwin, Gray, Chess, Clark, Berner, McCandlish, Radford, Sutskever, and Amodei]{brown2020language}
Tom Brown, Benjamin Mann, Nick Ryder, Melanie Subbiah, Jared~D Kaplan, Prafulla Dhariwal, Arvind Neelakantan, Pranav Shyam, Girish Sastry, Amanda Askell, Sandhini Agarwal, Ariel Herbert-Voss, Gretchen Krueger, Tom Henighan, Rewon Child, Aditya Ramesh, Daniel Ziegler, Jeffrey Wu, Clemens Winter, Chris Hesse, Mark Chen, Eric Sigler, Mateusz Litwin, Scott Gray, Benjamin Chess, Jack Clark, Christopher Berner, Sam McCandlish, Alec Radford, Ilya Sutskever, and Dario Amodei.
\newblock Language models are few-shot learners.
\newblock pp.\  1877--1901, 2020.

\bibitem[Cemri et~al.(2025)Cemri, Pan, Yang, Agrawal, Chopra, Tiwari, Keutzer, Parameswaran, Klein, Ramchandran, et~al.]{cemri2025multi}
Mert Cemri, Melissa~Z Pan, Shuyi Yang, Lakshya~A Agrawal, Bhavya Chopra, Rishabh Tiwari, Kurt Keutzer, Aditya Parameswaran, Dan Klein, Kannan Ramchandran, et~al.
\newblock Why do multi-agent llm systems fail?
\newblock \emph{arXiv preprint arXiv:2503.13657}, 2025.

\bibitem[Chaloner \& Verdinelli(1995)Chaloner and Verdinelli]{chaloner1995bayesian}
Kathryn Chaloner and Isabella Verdinelli.
\newblock {Bayesian Experimental Design: A Review}.
\newblock \emph{Statistical Science}, pp.\  273--304, 1995.

\bibitem[Chen et~al.(2024)Chen, Xu, Liang, He, Pang, Yu, Song, Liu, Zhou, Zhang, et~al.]{chen2024not}
Xingyu Chen, Jiahao Xu, Tian Liang, Zhiwei He, Jianhui Pang, Dian Yu, Linfeng Song, Qiuzhi Liu, Mengfei Zhou, Zhuosheng Zhang, et~al.
\newblock Do {N}ot {T}hink that {M}uch for 2+3=? {O}n the {O}verthinking of o1-like {LLM}s.
\newblock \emph{arXiv preprint arXiv:2412.21187}, 2024.

\bibitem[Chopra et~al.(2019)Chopra, Tessler, and Goodman]{chopra2019first}
Sahil Chopra, Michael~Henry Tessler, and Noah~D Goodman.
\newblock The first crank of the cultural ratchet: Learning and transmitting concepts through language.
\newblock In \emph{Proceedings of the {A}nnual {M}eeting of the {C}ognitive {S}cience {S}ociety}, volume~41, 2019.

\bibitem[Coda-Forno et~al.(2024)Coda-Forno, Binz, Wang, and Schulz]{coda2024cogbench}
Julian Coda-Forno, Marcel Binz, Jane~X Wang, and Eric Schulz.
\newblock Cogbench: {A} {L}arge {L}anguage {M}odel {W}alks into {A} {P}sychology {L}ab.
\newblock \emph{arXiv preprint arXiv:2402.18225}, 2024.

\bibitem[Cuadron et~al.(2025)Cuadron, Li, Ma, Wang, Wang, Zhuang, Liu, Schroeder, Xia, Mao, et~al.]{cuadron2025danger}
Alejandro Cuadron, Dacheng Li, Wenjie Ma, Xingyao Wang, Yichuan Wang, Siyuan Zhuang, Shu Liu, Luis~Gaspar Schroeder, Tian Xia, Huanzhi Mao, et~al.
\newblock The danger of overthinking: Examining the reasoning-action dilemma in agentic tasks.
\newblock \emph{arXiv preprint arXiv:2502.08235}, 2025.

\bibitem[DeGroot(1962)]{degroot1962uncertainty}
MH~DeGroot.
\newblock {Uncertainty, Information, and Sequential Experiments}.
\newblock \emph{The Annals of Mathematical Statistics}, 33\penalty0 (2):\penalty0 404--419, 1962.

\bibitem[Deisenroth \& Rasmussen(2011)Deisenroth and Rasmussen]{deisenroth2011pilco}
Marc Deisenroth and Carl~E Rasmussen.
\newblock {PILCO: A Model-Based and Data-Efficient Approach to Policy Search}.
\newblock In \emph{Proceedings of the 28th International Conference on Machine Learning (ICML-11)}, pp.\  465--472, 2011.

\bibitem[Der~Kiureghian \& Ditlevsen(2009)Der~Kiureghian and Ditlevsen]{der2009aleatory}
Armen Der~Kiureghian and Ove Ditlevsen.
\newblock {Aleatory or Epistemic? Does it Matter?}
\newblock \emph{Structural Safety}, 31\penalty0 (2):\penalty0 105--112, 2009.

\bibitem[Falck et~al.(2024)Falck, Wang, and Holmes]{falck2024context}
Fabian Falck, Ziyu Wang, and Chris Holmes.
\newblock {Is in-context learning in large language models Bayesian? A martingale perspective.}
\newblock \emph{arXiv preprint arXiv:2406.00793}, 2024.

\bibitem[Foster et~al.(2019)Foster, Jankowiak, Bingham, Horsfall, Teh, Rainforth, and Goodman]{foster2019variational}
Adam Foster, Martin Jankowiak, Elias Bingham, Paul Horsfall, Yee~Whye Teh, Thomas Rainforth, and Noah Goodman.
\newblock {Variational Bayesian optimal experimental design}.
\newblock \emph{Advances in Neural Information Processing Systems}, 32, 2019.

\bibitem[Frank(2023)]{frank2023baby}
Michael~C Frank.
\newblock Baby steps in evaluating the capacities of large language models.
\newblock \emph{Nature Reviews Psychology}, 2\penalty0 (8):\penalty0 451--452, 2023.

\bibitem[Gandhi et~al.(2025)Gandhi, Li, Goodyear, Li, Bhaskar, Zaman, and Goodman]{gandhi2025boxinggym}
Kanishk Gandhi, Michael~Y Li, Lyle Goodyear, Louise Li, Aditi Bhaskar, Mohammed Zaman, and Noah~D Goodman.
\newblock Boxing{G}ym: Benchmarking progress in automated experimental design and model discovery.
\newblock \emph{arXiv preprint arXiv:2501.01540}, 2025.

\bibitem[Gil et~al.(2014)Gil, Greaves, Hendler, and Hirsh]{gil2014amplify}
Yolanda Gil, Mark Greaves, James Hendler, and Haym Hirsh.
\newblock Amplify scientific discovery with artificial intelligence.
\newblock \emph{Science}, 346\penalty0 (6206):\penalty0 171--172, 2014.

\bibitem[Girden(1992)]{girden1992anova}
Ellen~R Girden.
\newblock \emph{ANOVA: Repeated measures}.
\newblock Number~84. Sage, 1992.

\bibitem[Gottweis et~al.(2025)Gottweis, Weng, Daryin, Tu, Palepu, Sirkovic, Myaskovsky, Weissenberger, Rong, Tanno, Saab, Popovici, Blum, Zhang, Chou, Hassidim, Gokturk, Vahdat, Kohli, Matias, Carroll, Kulkarni, Tomasev, Guan, Dhillon, Vaishnav, Lee, Costa, Penadés, Peltz, Xu, Pawlosky, Karthikesalingam, and Natarajan]{gottweis2025towards}
Juraj Gottweis, Wei-Hung Weng, Alexander Daryin, Tao Tu, Anil Palepu, Petar Sirkovic, Artiom Myaskovsky, Felix Weissenberger, Keran Rong, Ryutaro Tanno, Khaled Saab, Dan Popovici, Jacob Blum, Fan Zhang, Katherine Chou, Avinatan Hassidim, Burak Gokturk, Amin Vahdat, Pushmeet Kohli, Yossi Matias, Andrew Carroll, Kavita Kulkarni, Nenad Tomasev, Yuan Guan, Vikram Dhillon, Eeshit~Dhaval Vaishnav, Byron Lee, Tiago R~D Costa, José~R Penadés, Gary Peltz, Yunhan Xu, Annalisa Pawlosky, Alan Karthikesalingam, and Vivek Natarajan.
\newblock Towards an {AI} co-scientist.
\newblock \emph{arXiv preprint arXiv:2502.18864}, 2025.

\bibitem[Grattafiori et~al.(2024)Grattafiori, Dubey, Jauhri, Pandey, Kadian, Al-Dahle, Letman, Mathur, Schelten, Vaughan, Yang, Fan, Goyal, Hartshorn, Yang, Mitra, Sravankumar, Korenev, Hinsvark, Rao, Zhang, Rodriguez, Gregerson, Spataru, Roziere, Biron, Tang, Chern, Caucheteux, Nayak, Bi, Marra, McConnell, Keller, Touret, Wu, Wong, Ferrer, Nikolaidis, Allonsius, Song, Pintz, Livshits, Wyatt, Esiobu, Choudhary, Mahajan, Garcia-Olano, Perino, Hupkes, Lakomkin, AlBadawy, Lobanova, Dinan, Smith, Radenovic, Guzmán, Zhang, Synnaeve, Lee, Anderson, Thattai, Nail, Mialon, Pang, Cucurell, Nguyen, Korevaar, Xu, Touvron, Zarov, Ibarra, Kloumann, Misra, Evtimov, Zhang, Copet, Lee, Geffert, Vranes, Park, Mahadeokar, Shah, van~der Linde, Billock, Hong, Lee, Fu, Chi, Huang, Liu, Wang, Yu, Bitton, Spisak, Park, Rocca, Johnstun, Saxe, Jia, Alwala, Prasad, Upasani, Plawiak, Li, Heafield, Stone, El-Arini, Iyer, Malik, Chiu, Bhalla, Lakhotia, Rantala-Yeary, van~der Maaten, Chen, Tan, Jenkins, Martin, Madaan, Malo, Blecher,
  Landzaat, de~Oliveira, Muzzi, Pasupuleti, Singh, Paluri, Kardas, Tsimpoukelli, Oldham, Rita, Pavlova, Kambadur, Lewis, Si, Singh, Hassan, Goyal, Torabi, Bashlykov, Bogoychev, Chatterji, Zhang, Duchenne, Çelebi, Alrassy, Zhang, Li, Vasic, Weng, Bhargava, Dubal, Krishnan, Koura, Xu, He, Dong, Srinivasan, Ganapathy, Calderer, Cabral, Stojnic, Raileanu, Maheswari, Girdhar, Patel, Sauvestre, Polidoro, Sumbaly, Taylor, Silva, Hou, Wang, Hosseini, Chennabasappa, Singh, Bell, Kim, Edunov, Nie, Narang, Raparthy, Shen, Wan, Bhosale, Zhang, Vandenhende, Batra, Whitman, Sootla, Collot, Gururangan, Borodinsky, Herman, Fowler, Sheasha, Georgiou, Scialom, Speckbacher, Mihaylov, Xiao, Karn, Goswami, Gupta, Ramanathan, Kerkez, Gonguet, Do, Vogeti, Albiero, Petrovic, Chu, Xiong, Fu, Meers, Martinet, Wang, Wang, Tan, Xia, Xie, Jia, Wang, Goldschlag, Gaur, Babaei, Wen, Song, Zhang, Li, Mao, Coudert, Yan, Chen, and Papakipos]{grattafiori2024llama3herd}
Aaron Grattafiori, Abhimanyu Dubey, Abhinav Jauhri, Abhinav Pandey, Abhishek Kadian, Ahmad Al-Dahle, Aiesha Letman, Akhil Mathur, Alan Schelten, Alex Vaughan, Amy Yang, Angela Fan, Anirudh Goyal, Anthony Hartshorn, Aobo Yang, Archi Mitra, Archie Sravankumar, Artem Korenev, Arthur Hinsvark, Arun Rao, Aston Zhang, Aurelien Rodriguez, Austen Gregerson, Ava Spataru, Baptiste Roziere, Bethany Biron, Binh Tang, Bobbie Chern, Charlotte Caucheteux, Chaya Nayak, Chloe Bi, Chris Marra, Chris McConnell, Christian Keller, Christophe Touret, Chunyang Wu, Corinne Wong, Cristian~Canton Ferrer, Cyrus Nikolaidis, Damien Allonsius, Daniel Song, Danielle Pintz, Danny Livshits, Danny Wyatt, David Esiobu, Dhruv Choudhary, Dhruv Mahajan, Diego Garcia-Olano, Diego Perino, Dieuwke Hupkes, Egor Lakomkin, Ehab AlBadawy, Elina Lobanova, Emily Dinan, Eric~Michael Smith, Filip Radenovic, Francisco Guzmán, Frank Zhang, Gabriel Synnaeve, Gabrielle Lee, Georgia~Lewis Anderson, Govind Thattai, Graeme Nail, Gregoire Mialon, Guan Pang,
  Guillem Cucurell, Hailey Nguyen, Hannah Korevaar, Hu~Xu, Hugo Touvron, Iliyan Zarov, Imanol~Arrieta Ibarra, Isabel Kloumann, Ishan Misra, Ivan Evtimov, Jack Zhang, Jade Copet, Jaewon Lee, Jan Geffert, Jana Vranes, Jason Park, Jay Mahadeokar, Jeet Shah, Jelmer van~der Linde, Jennifer Billock, Jenny Hong, Jenya Lee, Jeremy Fu, Jianfeng Chi, Jianyu Huang, Jiawen Liu, Jie Wang, Jiecao Yu, Joanna Bitton, Joe Spisak, Jongsoo Park, Joseph Rocca, Joshua Johnstun, Joshua Saxe, Junteng Jia, Kalyan~Vasuden Alwala, Karthik Prasad, Kartikeya Upasani, Kate Plawiak, Ke~Li, Kenneth Heafield, Kevin Stone, Khalid El-Arini, Krithika Iyer, Kshitiz Malik, Kuenley Chiu, Kunal Bhalla, Kushal Lakhotia, Lauren Rantala-Yeary, Laurens van~der Maaten, Lawrence Chen, Liang Tan, Liz Jenkins, Louis Martin, Lovish Madaan, Lubo Malo, Lukas Blecher, Lukas Landzaat, Luke de~Oliveira, Madeline Muzzi, Mahesh Pasupuleti, Mannat Singh, Manohar Paluri, Marcin Kardas, Maria Tsimpoukelli, Mathew Oldham, Mathieu Rita, Maya Pavlova, Melanie Kambadur,
  Mike Lewis, Min Si, Mitesh~Kumar Singh, Mona Hassan, Naman Goyal, Narjes Torabi, Nikolay Bashlykov, Nikolay Bogoychev, Niladri Chatterji, Ning Zhang, Olivier Duchenne, Onur Çelebi, Patrick Alrassy, Pengchuan Zhang, Pengwei Li, Petar Vasic, Peter Weng, Prajjwal Bhargava, Pratik Dubal, Praveen Krishnan, Punit~Singh Koura, Puxin Xu, Qing He, Qingxiao Dong, Ragavan Srinivasan, Raj Ganapathy, Ramon Calderer, Ricardo~Silveira Cabral, Robert Stojnic, Roberta Raileanu, Rohan Maheswari, Rohit Girdhar, Rohit Patel, Romain Sauvestre, Ronnie Polidoro, Roshan Sumbaly, Ross Taylor, Ruan Silva, Rui Hou, Rui Wang, Saghar Hosseini, Sahana Chennabasappa, Sanjay Singh, Sean Bell, Seohyun~Sonia Kim, Sergey Edunov, Shaoliang Nie, Sharan Narang, Sharath Raparthy, Sheng Shen, Shengye Wan, Shruti Bhosale, Shun Zhang, Simon Vandenhende, Soumya Batra, Spencer Whitman, Sten Sootla, Stephane Collot, Suchin Gururangan, Sydney Borodinsky, Tamar Herman, Tara Fowler, Tarek Sheasha, Thomas Georgiou, Thomas Scialom, Tobias Speckbacher,
  Todor Mihaylov, Tong Xiao, Ujjwal Karn, Vedanuj Goswami, Vibhor Gupta, Vignesh Ramanathan, Viktor Kerkez, Vincent Gonguet, Virginie Do, Vish Vogeti, Vítor Albiero, Vladan Petrovic, Weiwei Chu, Wenhan Xiong, Wenyin Fu, Whitney Meers, Xavier Martinet, Xiaodong Wang, Xiaofang Wang, Xiaoqing~Ellen Tan, Xide Xia, Xinfeng Xie, Xuchao Jia, Xuewei Wang, Yaelle Goldschlag, Yashesh Gaur, Yasmine Babaei, Yi~Wen, Yiwen Song, Yuchen Zhang, Yue Li, Yuning Mao, Zacharie~Delpierre Coudert, Zheng Yan, Zhengxing Chen, and Zoe Papakipos.
\newblock The llama 3 herd of models, 2024.
\newblock URL \url{https://arxiv.org/abs/2407.21783}.
\newblock Accessed: 2025-05-14.

\bibitem[Griffiths et~al.(2024)Griffiths, Zhu, Grant, and Thomas~McCoy]{griffiths2024bayes}
Thomas~L Griffiths, Jian-Qiao Zhu, Erin Grant, and R~Thomas~McCoy.
\newblock {Bayes in the Age of Intelligent Machine}s.
\newblock \emph{Current Directions in Psychological Science}, 33\penalty0 (5):\penalty0 283--291, 2024.

\bibitem[Guo et~al.(2025)Guo, Yang, Zhang, Song, Zhang, Xu, Zhu, Ma, Wang, Bi, et~al.]{guo2025deepseek}
Daya Guo, Dejian Yang, Haowei Zhang, Junxiao Song, Ruoyu Zhang, Runxin Xu, Qihao Zhu, Shirong Ma, Peiyi Wang, Xiao Bi, et~al.
\newblock Deepseek-r1: Incentivizing reasoning capability in llms via reinforcement learning.
\newblock \emph{arXiv preprint arXiv:2501.12948}, 2025.

\bibitem[He et~al.(2025)He, Li, Liu, Wang, Bu, Zhang, Peng, Zhang, Su, and Zheng]{he2025can}
Yancheng He, Shilong Li, Jiaheng Liu, Weixun Wang, Xingyuan Bu, Ge~Zhang, Zhongyuan Peng, Zhaoxiang Zhang, Wenbo Su, and Bo~Zheng.
\newblock Can large language models detect errors in long chain-of-thought reasoning?
\newblock \emph{arXiv preprint arXiv:2502.19361}, 2025.

\bibitem[Hurst et~al.(2024)Hurst, Lerer, Goucher, Perelman, Ramesh, Clark, Ostrow, Welihinda, Hayes, Radford, et~al.]{hurst2024gpt}
Aaron Hurst, Adam Lerer, Adam~P Goucher, Adam Perelman, Aditya Ramesh, Aidan Clark, AJ~Ostrow, Akila Welihinda, Alan Hayes, Alec Radford, et~al.
\newblock {GPT}-4o system card.
\newblock \emph{arXiv preprint arXiv:2410.21276}, 2024.

\bibitem[Kang et~al.(2024)Kang, Setlur, Ghosh, Steinhardt, Tomlin, Levine, and Kumar]{kang2024learning}
Katie Kang, Amrith Setlur, Dibya Ghosh, Jacob Steinhardt, Claire Tomlin, Sergey Levine, and Aviral Kumar.
\newblock What do learning dynamics reveal about generalization in llm reasoning?
\newblock \emph{arXiv preprint arXiv:2411.07681}, 2024.

\bibitem[Ku et~al.(2025)Ku, Campbell, Bai, Geng, Liu, Marjieh, McCoy, Nam, Sucholutsky, Veselovsky, et~al.]{ku2025using}
Alexander Ku, Declan Campbell, Xuechunzi Bai, Jiayi Geng, Ryan Liu, Raja Marjieh, R~Thomas McCoy, Andrew Nam, Ilia Sucholutsky, Veniamin Veselovsky, et~al.
\newblock Using the tools of cognitive science to understand large language models at different levels of analysis.
\newblock \emph{arXiv preprint arXiv:2503.13401}, 2025.

\bibitem[La~Torre et~al.(2007)La~Torre, Madhusudan, and Parlato]{la2007robust}
Salvatore La~Torre, Parthasarathy Madhusudan, and Gennaro Parlato.
\newblock A robust class of context-sensitive languages.
\newblock In \emph{22nd Annual IEEE {S}ymposium on {L}ogic in {C}omputer {S}cience (LICS 2007)}, pp.\  161--170, 2007.

\bibitem[Lampinen et~al.(2023)Lampinen, Chan, Dasgupta, Nam, and Wang]{lampinen2023passive}
Andrew~Kyle Lampinen, Stephanie C~Y Chan, Ishita Dasgupta, Andrew~J Nam, and Jane~X Wang.
\newblock Passive learning of active causal strategies in agents and language models.
\newblock \emph{Advances in Neural Information Processing Systems}, 2023.

\bibitem[Li et~al.(2025)Li, Li, Li, and Jin]{li2025structured}
Jia Li, Ge~Li, Yongmin Li, and Zhi Jin.
\newblock Structured chain-of-thought prompting for code generation.
\newblock \emph{ACM Transactions on Software Engineering and Methodology}, 34\penalty0 (2):\penalty0 1--23, 2025.

\bibitem[Li \& Littman(2010)Li and Littman]{li2010reducing}
Lihong Li and Michael~L Littman.
\newblock {Reducing Reinforcement Learning to KWIK Online Regression}.
\newblock \emph{Annals of Mathematics and Artificial Intelligence}, 58:\penalty0 217--237, 2010.

\bibitem[Li et~al.(2008)Li, Littman, and Walsh]{li2008knows}
Lihong Li, Michael~L Littman, and Thomas~J Walsh.
\newblock {Knows What It Knows: A Framework for Self-Aware Learning}.
\newblock In \emph{Proceedings of the 25th International Conference on Machine Learning}, pp.\  568--575, 2008.

\bibitem[Lindley(1956)]{lindley1956measure}
Dennis~V Lindley.
\newblock On a measure of the information provided by an experiment.
\newblock \emph{The Annals of Mathematical Statistics}, 27\penalty0 (4):\penalty0 986--1005, 1956.

\bibitem[Littlestone(1988)]{littlestone1988learning}
Nick Littlestone.
\newblock Learning quickly when irrelevant attributes abound: A new linear-threshold algorithm.
\newblock \emph{Machine Learning}, 2:\penalty0 285--318, 1988.

\bibitem[Liu et~al.(2021)Liu, Raghunathan, Liang, and Finn]{liu2021decoupling}
Evan~Z Liu, Aditi Raghunathan, Percy Liang, and Chelsea Finn.
\newblock {Decoupling Exploration and Exploitation for Meta-Reinforcement Learning Without Sacrifices}.
\newblock In \emph{International Conference on Machine Learning}, pp.\  6925--6935, 2021.

\bibitem[Liu et~al.(2024{\natexlab{a}})Liu, Geng, Peterson, Sucholutsky, and Griffiths]{liu2024large}
Ryan Liu, Jiayi Geng, Joshua~C Peterson, Ilia Sucholutsky, and Thomas~L Griffiths.
\newblock Large language models assume people are more rational than we really are.
\newblock \emph{arXiv preprint arXiv:2406.17055}, 2024{\natexlab{a}}.

\bibitem[Liu et~al.(2024{\natexlab{b}})Liu, Geng, Wu, Sucholutsky, Lombrozo, and Griffiths]{liu2024mind}
Ryan Liu, Jiayi Geng, Addison~J Wu, Ilia Sucholutsky, Tania Lombrozo, and Thomas~L Griffiths.
\newblock Mind your step (by step): Chain-of-thought can reduce performance on tasks where thinking makes humans worse.
\newblock \emph{arXiv preprint arXiv:2410.21333}, 2024{\natexlab{b}}.

\bibitem[Lu et~al.(2024)Lu, Lu, Lange, Foerster, Clune, and Ha]{lu2024the}
Chris Lu, Cong Lu, Robert~Tjarko Lange, Jakob Foerster, Jeff Clune, and David Ha.
\newblock The ai scientist: Towards fully automated open-ended scientific discovery.
\newblock \emph{arXiv preprint arXiv:2408.06292}, 2024.

\bibitem[Luo et~al.(2025)Luo, Song, Zhang, Liu, Wang, Chen, Su, and Zheng]{luo2025deconstructing}
Yijia Luo, Yulin Song, Xingyao Zhang, Jiaheng Liu, Weixun Wang, GengRu Chen, Wenbo Su, and Bo~Zheng.
\newblock Deconstructing long chain-of-thought: A structured reasoning optimization framework for long cot distillation.
\newblock \emph{arXiv preprint arXiv:2503.16385}, 2025.

\bibitem[Markant \& Gureckis(2010)Markant and Gureckis]{markant2010category}
Doug Markant and Todd Gureckis.
\newblock Category learning through active sampling.
\newblock In \emph{Proceedings of the {A}nnual {M}eeting of the {C}ognitive {S}cience {S}ociety}, volume~32, 2010.

\bibitem[Markant \& Gureckis(2014)Markant and Gureckis]{markant2014better}
Douglas~B Markant and Todd~M Gureckis.
\newblock Is it better to select or to receive? learning via active and passive hypothesis testing.
\newblock \emph{Journal of Experimental Psychology: General}, 143\penalty0 (1):\penalty0 94, 2014.

\bibitem[McCoy \& Griffiths(2023)McCoy and Griffiths]{mccoy2023modeling}
R~Thomas McCoy and Thomas~L Griffiths.
\newblock Modeling rapid language learning by distilling bayesian priors into artificial neural networks.
\newblock \emph{arXiv preprint arXiv:2305.14701}, 2023.

\bibitem[McCoy et~al.(2024)McCoy, Yao, Friedman, Hardy, and Griffiths]{mccoy2024embers}
R~Thomas McCoy, Shunyu Yao, Dan Friedman, Mathew~D Hardy, and Thomas~L Griffiths.
\newblock Embers of autoregression show how large language models are shaped by the problem they are trained to solve.
\newblock \emph{Proceedings of the National Academy of Sciences}, 121\penalty0 (41):\penalty0 e2322420121, 2024.

\bibitem[Ni et~al.(2024)Ni, Li, Hu, Han, Xu, Chen, Liu, Ye, and Bai]{ni2024matpilot}
Ziqi Ni, Yahao Li, Kaijia Hu, Kunyuan Han, Ming Xu, Xingyu Chen, Fengqi Liu, Yicong Ye, and Shuxin Bai.
\newblock Matpilot: an llm-enabled ai materials scientist under the framework of human-machine collaboration.
\newblock \emph{arXiv preprint arXiv:2411.08063}, 2024.

\bibitem[O'Neill et~al.(2025)O'Neill, Ghosal, R{\u{a}}ileanu, Walmsley, Bui, Schawinski, and Ciuc{\u{a}}]{o2025sparks}
Charles O'Neill, Tirthankar Ghosal, Roberta R{\u{a}}ileanu, Mike Walmsley, Thang Bui, Kevin Schawinski, and Ioana Ciuc{\u{a}}.
\newblock Sparks of science: Hypothesis generation using structured paper data.
\newblock \emph{arXiv preprint arXiv:2504.12976}, 2025.

\bibitem[Osband et~al.(2013)Osband, Russo, and Van~Roy]{osband2013more}
Ian Osband, Daniel Russo, and Benjamin Van~Roy.
\newblock {(More) Efficient Reinforcement Learning via Posterior Sampling}.
\newblock \emph{Advances in Neural Information Processing Systems}, 26:\penalty0 3003--3011, 2013.

\bibitem[Puterman(1994)]{Puterman94}
Martin~L. Puterman.
\newblock \emph{{Markov Decision Processes---Discrete Stochastic Dynamic Programming}}.
\newblock John Wiley \& Sons, 1994.

\bibitem[Reddy \& Shojaee(2025)Reddy and Shojaee]{reddy2025towards}
Chandan~K Reddy and Parshin Shojaee.
\newblock Towards scientific discovery with generative ai: Progress, opportunities, and challenges.
\newblock In \emph{Proceedings of the AAAI Conference on Artificial Intelligence}, volume~39, pp.\  28601--28609, 2025.

\bibitem[Rivest \& Schapire(1987)Rivest and Schapire]{rivest1987diversity}
Ronald~L Rivest and Robert~E Schapire.
\newblock {Diversity-Based Inference of Finite Automata}.
\newblock In \emph{28th Annual Symposium on Foundations of Computer Science (SFCS 1987)}, pp.\  78--87, 1987.

\bibitem[Rivest \& Schapire(1989)Rivest and Schapire]{rivest1989inference}
Ronald~L Rivest and Robert~E Schapire.
\newblock {Inference of Finite Automata Using Homing Sequences}.
\newblock In \emph{Proceedings of the Twenty-First Annual ACM Symposium on Theory of Computing}, pp.\  411--420, 1989.

\bibitem[Rmus et~al.(2025)Rmus, Jagadish, Mathony, Ludwig, and Schulz]{rmus2025towards}
Milena Rmus, Akshay~K. Jagadish, Marvin Mathony, Tobias Ludwig, and Eric Schulz.
\newblock Towards automation of cognitive modeling using large language models.
\newblock \emph{arXiv preprint arXiv:2502.00879}, 2025.

\bibitem[Rule et~al.(2024)Rule, Piantadosi, Cropper, Ellis, Nye, and Tenenbaum]{rule2024symbolic}
Joshua~S Rule, Steven~T Piantadosi, Andrew Cropper, Kevin Ellis, Maxwell Nye, and Joshua~B Tenenbaum.
\newblock Symbolic metaprogram search improves learning efficiency and explains rule learning in humans.
\newblock \emph{Nature Communications}, 15\penalty0 (1):\penalty0 6847, 2024.

\bibitem[Sayedi et~al.(2010)Sayedi, Zadimoghaddam, and Blum]{sayedi2010trading}
Amin Sayedi, Morteza Zadimoghaddam, and Avrim Blum.
\newblock {Trading off Mistakes and Don't-Know Predictions}.
\newblock \emph{Advances in Neural Information Processing Systems}, 23, 2010.

\bibitem[Schmidgall et~al.(2025)Schmidgall, Su, Wang, Sun, Wu, Yu, Liu, Liu, and Barsoum]{schmidgall2025agent}
Samuel Schmidgall, Yusheng Su, Ze~Wang, Ximeng Sun, Jialian Wu, Xiaodong Yu, Jiang Liu, Zicheng Liu, and Emad Barsoum.
\newblock Agent laboratory: Using llm agents as research assistants.
\newblock \emph{arXiv preprint arXiv:2501.04227}, 2025.

\bibitem[Settles(2009)]{settlestr09}
Burr Settles.
\newblock {Active Learning Literature Survey}.
\newblock Computer Sciences Technical Report 1648, University of Wisconsin--Madison, 2009.

\bibitem[Shojaee et~al.(2025)Shojaee, Meidani, Gupta, Farimani, and Reddy]{shojaee2025llm}
Parshin Shojaee, Kazem Meidani, Shashank Gupta, Amir~Barati Farimani, and Chandan~K Reddy.
\newblock Llm-sr: Scientific equation discovery via programming with large language models.
\newblock \emph{International Conference on Learning Representations}, 2025.

\bibitem[Si et~al.(2024)Si, Yang, and Hashimoto]{si2024can}
Chenglei Si, Diyi Yang, and Tatsunori Hashimoto.
\newblock {Can LLMs generate novel research ideas? A large-scale human study with 100+ NLP researchers}.
\newblock \emph{arXiv preprint arXiv:2409.04109}, 2024.

\bibitem[Sprague et~al.(2024)Sprague, Yin, Rodriguez, Jiang, Wadhwa, Singhal, Zhao, Ye, Mahowald, and Durrett]{sprague2024cot}
Zayne Sprague, Fangcong Yin, Juan~Diego Rodriguez, Dongwei Jiang, Manya Wadhwa, Prasann Singhal, Xinyu Zhao, Xi~Ye, Kyle Mahowald, and Greg Durrett.
\newblock {To CoT or not to CoT? chain-of-thought helps mainly on math and symbolic reasoning}.
\newblock \emph{arXiv preprint arXiv:2409.12183}, 2024.

\bibitem[Starace et~al.(2025)Starace, Jaffe, Sherburn, Aung, Chan, Maksin, Dias, Mays, Kinsella, Thompson, et~al.]{starace2025paperbench}
Giulio Starace, Oliver Jaffe, Dane Sherburn, James Aung, Jun~Shern Chan, Leon Maksin, Rachel Dias, Evan Mays, Benjamin Kinsella, Wyatt Thompson, et~al.
\newblock Paperbench: Evaluating ai's ability to replicate ai research.
\newblock \emph{arXiv preprint arXiv:2504.01848}, 2025.

\bibitem[Strehl \& Littman(2008)Strehl and Littman]{strehl2008analysis}
Alexander~L Strehl and Michael~L Littman.
\newblock {An Analysis of Model-based interval estimation for Markov Decision Processes}.
\newblock \emph{Journal of Computer and System Sciences}, 74\penalty0 (8):\penalty0 1309--1331, 2008.

\bibitem[Strens(2000)]{strens2000bayesian}
Malcolm~JA Strens.
\newblock A {B}ayesian framework for reinforcement learning.
\newblock In \emph{Proceedings of the Seventeenth International Conference on Machine Learning}, pp.\  943--950, 2000.

\bibitem[Sui et~al.(2025)Sui, Chuang, Wang, Zhang, Zhang, Yuan, Liu, Wen, Chen, Hu, et~al.]{sui2025stop}
Yang Sui, Yu-Neng Chuang, Guanchu Wang, Jiamu Zhang, Tianyi Zhang, Jiayi Yuan, Hongyi Liu, Andrew Wen, Hanjie Chen, Xia Hu, et~al.
\newblock Stop overthinking: A survey on efficient reasoning for large language models.
\newblock \emph{arXiv preprint arXiv:2503.16419}, 2025.

\bibitem[Sutton(1990)]{sutton1990integrated}
Richard~S Sutton.
\newblock Integrated architectures for learning, planning, and reacting based on approximating dynamic programming.
\newblock In \emph{Proceedings of the Seventh International Conference on Machine Learning}, pp.\  216--224, 1990.

\bibitem[Sutton(1991)]{sutton1991dyna}
Richard~S Sutton.
\newblock Dyna, an integrated architecture for learning, planning, and reacting.
\newblock \emph{ACM Sigart Bulletin}, 2\penalty0 (4):\penalty0 160--163, 1991.

\bibitem[Sutton \& Barto(1998)Sutton and Barto]{sutton1998introduction}
Richard~S Sutton and Andrew~G Barto.
\newblock \emph{{Introduction to Reinforcement Learning}}.
\newblock MIT Press, 1998.

\bibitem[Szita \& Szepesv{\'a}ri(2011)Szita and Szepesv{\'a}ri]{szita2011agnostic}
Istv{\'a}n Szita and Csaba Szepesv{\'a}ri.
\newblock {Agnostic KWIK learning and Efficient Approximate Reinforcement Learning}.
\newblock In \emph{Proceedings of the 24th Annual Conference on Learning Theory}, pp.\  739--772, 2011.

\bibitem[Thrun \& M{\"o}ller(1991)Thrun and M{\"o}ller]{thrun1991active}
Sebastian~B Thrun and Knut M{\"o}ller.
\newblock Active exploration in dynamic environments.
\newblock \emph{Advances in Neural Information Processing Systems}, 4, 1991.

\bibitem[Walsh et~al.(2009)Walsh, Szita, Diuk, and Littman]{walsh2009exploring}
Thomas~J Walsh, Istv{\'a}n Szita, Carlos Diuk, and Michael~L Littman.
\newblock {Exploring Compact Reinforcement-Learning Representations with Linear Regression}.
\newblock In \emph{Proceedings of the Twenty-Fifth Conference on Uncertainty in Artificial Intelligence}, pp.\  591--598, 2009.

\bibitem[Wang et~al.(2023)Wang, Fu, Du, Gao, Huang, Liu, Chandak, Liu, Van~Katwyk, Deac, et~al.]{wang2023scientific}
Hanchen Wang, Tianfan Fu, Yuanqi Du, Wenhao Gao, Kexin Huang, Ziming Liu, Payal Chandak, Shengchao Liu, Peter Van~Katwyk, Andreea Deac, et~al.
\newblock {Scientific Discovery in the Age of Artificial Intelligence}.
\newblock \emph{Nature}, 620\penalty0 (7972):\penalty0 47--60, 2023.

\bibitem[Wang et~al.(2025)Wang, Liu, Xu, Liang, Chen, He, Song, Yu, Li, Zhang, et~al.]{wang2025thoughts}
Yue Wang, Qiuzhi Liu, Jiahao Xu, Tian Liang, Xingyu Chen, Zhiwei He, Linfeng Song, Dian Yu, Juntao Li, Zhuosheng Zhang, et~al.
\newblock Thoughts are all over the place: On the underthinking of o1-like llms.
\newblock \emph{arXiv preprint arXiv:2501.18585}, 2025.

\bibitem[Wei et~al.(2022)Wei, Wang, Schuurmans, Bosma, Xia, Chi, Le, Zhou, et~al.]{wei2022chain}
Jason Wei, Xuezhi Wang, Dale Schuurmans, Maarten Bosma, Fei Xia, Ed~Chi, Quoc~V Le, Denny Zhou, et~al.
\newblock Chain-of-thought prompting elicits reasoning in large language models.
\newblock \emph{Advances in {N}eural {I}nformation {P}rocessing {S}ystems}, 35:\penalty0 24824--24837, 2022.

\bibitem[Xie et~al.(2021)Xie, Raghunathan, Liang, and Ma]{xie2021explanation}
Sang~Michael Xie, Aditi Raghunathan, Percy Liang, and Tengyu Ma.
\newblock An explanation of in-context learning as implicit {B}ayesian inference.
\newblock In \emph{International Conference on Learning Representations}, 2021.

\bibitem[Yang \& Piantadosi(2022)Yang and Piantadosi]{yang2022}
Yuan Yang and Steven~T. Piantadosi.
\newblock One model for the learning of language.
\newblock \emph{Proceedings of the National Academy of Sciences}, 119\penalty0 (5):\penalty0 e2021865119, 2022.

\bibitem[Ying et~al.(2025)Ying, Collins, Wong, Sucholutsky, Liu, Weller, Shu, Griffiths, and Tenenbaum]{ying2025benchmarking}
Lance Ying, Katherine~M Collins, Lionel Wong, Ilia Sucholutsky, Ryan Liu, Adrian Weller, Tianmin Shu, Thomas~L Griffiths, and Joshua~B Tenenbaum.
\newblock On benchmarking human-like intelligence in machines.
\newblock \emph{arXiv preprint arXiv:2502.20502}, 2025.

\bibitem[Zhang et~al.(2024)Zhang, Collins, Wong, Barba, Weller, and Tenenbaum]{zhang2024people}
Cedegao~E Zhang, Katherine~M Collins, Lionel Wong, Mauricio Barba, Adrian Weller, and Joshua~B Tenenbaum.
\newblock People use fast, goal-directed simulation to reason about novel games.
\newblock \emph{arXiv preprint arXiv:2407.14095}, 2024.

\bibitem[Zhu \& Griffiths(2024{\natexlab{a}})Zhu and Griffiths]{zhu2024eliciting}
Jian-Qiao Zhu and Thomas~L Griffiths.
\newblock Eliciting the priors of large language models using iterated in-context learning.
\newblock \emph{arXiv preprint arXiv:2406.01860}, 2024{\natexlab{a}}.

\bibitem[Zhu \& Griffiths(2024{\natexlab{b}})Zhu and Griffiths]{zhu2024incoherent}
Jian-Qiao Zhu and Thomas~L Griffiths.
\newblock Incoherent probability judgments in large language models.
\newblock \emph{arXiv preprint arXiv:2401.16646}, 2024{\natexlab{b}}.

\end{thebibliography}
\bibliographystyle{neurips_2025}
\clearpage
\appendix
\label{sec:appendix}

\section{Black Box Designs}
\label{sec:blackbox_design}

\paragraph{Program} 
We used 100 list-mapping program instances from \citep{rule2024symbolic} to design the Program black-box API. Each black-box instance represents as a symbolic program defined in a domain-specific language (DSL). We implemented an interpreter pipeline that parses DSL expressions into abstract syntax trees and compiles them into executable Python code.

Each black-box supports two modes: \texttt{observation} (observation-only) and \texttt{intervention} (observation-intervention). In the \texttt{observation} mode, the black-box takes a random input list and returns the output produced by the underlying symbolic program, generating paired observational data:
\[
\text{input list} \rightarrow \text{program execution} \rightarrow \text{output list}
\]

In the \texttt{intervention} mode, the LLM queries an input or explicitly specifies an input-output pair. The black-box generates the output list or evaluates whether the given output matches the internally computed output and provides clear feedback:
\[
\text{Feedback} = \begin{cases}
\text{"output $\Rightarrow$ Correct"}, & \text{if the provided output matches the program output,}\\[6pt]
\text{"output $\Rightarrow$ Incorrect"}, & \text{otherwise.}
\end{cases}
\]

\paragraph{Formal Language}  
We followed \citep{yang2022,mccoy2023modeling} to implement a collection of 46 formal language instances to construct our formal language black-box, each instance being capable of generating strings according to specific symbolic rules ($e.g.$ $A^nB^n$). Each black-box instance behaves as an API from a generative model, operating in two modes: \texttt{observation} and \texttt{intervention}. 

In the \texttt{observation} mode (observation-only), the black-box randomly produces valid strings from its underlying rule, explicitly labeling each as generated output, for example:
\[
\text{``AAAABBBB'' is generated by the black-box.}
\]

In the \texttt{intervention} mode (observation-intervention), the LLM submits a specific string query to the black-box, which evaluates whether the string complies with its rule. The black-box responds clearly, indicating either acceptance or rejection:
\[
\text{Response} = 
\begin{cases}
\text{``[string] is generated by the black-box''}, & \text{if the strings compile with the rule,}\\[6pt]
\text{``[string] cannot be generated by the black-box''}, & \text{otherwise.}
\end{cases}
\]

To avoid generating infinite strings, we imposed a maximum character length of 64 for all single characters generated by the black-box.

\paragraph{Math Equation} 

For the math equation, we implemented the CES utility model as the black-box, designing it as a generative model capable of generating observational data or responding to queries from an LLM. The utility function of CES is mathematically defined as:
\[
U = \left(\sum_i a_i x_i^r\right)^{\frac{1}{r}},
\]
where the weights \(a_i\) satisfy the constraint \(\sum_i a_i = 1\), the parameter \(r\) controls the substitution elasticity, and \(x_i\) represents the quantities of goods in a basket.

CES black-box also provides two operational modes: \texttt{observation} (observation-only) and \texttt{intervention} (observation-intervention). In the \texttt{observation} mode, the black-box randomly samples two baskets (each a list of good quantities) and computes their utilities using the CES formulation. It then returns the preference outcome indicating which basket is preferred based on higher utility:
\[
\text{Preference} = 
\begin{cases}
\text{Basket1}, & U(\text{Basket1}) > U(\text{Basket2}), \\[6pt]
\text{Basket2}, & U(\text{Basket1}) < U(\text{Basket2}), \\[6pt]
\text{equal utility}, & U(\text{Basket1}) = U(\text{Basket2}).
\end{cases}
\]

In the \texttt{intervention} mode, an external model explicitly queries the black-box by specifying two baskets. In addition, the external model can also provide an estimated preference. The CES black-box internally evaluates the utilities based on the specified baskets and returns the actual preference outcome or feedback indicating whether the provided estimate was ``correct'' or ``incorrect''.

\section{Bayesian models as the `Optimal' Reference}

We employ Bayesian models as an oracle for optimal reverse-engineering against which we may assess the capabilities of LLMs.
Unlike LLMs, Bayesian models explicitly perform probabilistic inference within a clearly defined hypothesis space, systematically updating posterior beliefs using the Bayes rule to identify the underlying mechanism that best explain observed data.
Under the critical assumption that the true underlying rule resides within this hypothesis space (that is, the standard assumption of a well-specified prior), Bayesian models serve as an optimal reference standard in our experimental setting. We hypothesize that LLMs, when provided only with passive observational data, are unable to effectively utilize available information due to their inherent reliance on prior knowledge, resulting in significantly lower performance compared to the Bayesian optimal standard. However, allowing LLMs to actively intervene and collect data can substantially reduce the performance gap. For each of the three black-box systems evaluated, we replicated the Bayesian models from their original studies, adapting them to closely match our experimental conditions. Specifically, we provide Bayesian models with observed data generated by our black-box systems as an ideal reference. We also provide Bayesian models with the actively collected data from LLMs intervention to assess the informativeness of the data gathered by LLMs. To ensure rigorous comparability, we applied identical evaluation methodologies to both the Bayesian models and LLMs.


\paragraph{Program} 
We used the Bayesian inference approach from \cite{rule2024symbolic} to establish an optimal reference for list-mapping program black-box. Specifically, we utilized their MetaProgram Learner, which performs Bayesian inference over symbolic metaprograms that generate target programs from observed data.

Given observational data \( D \), consisting of input-output pairs generated by symbolic programs, the MPL computes the posterior distribution over candidate hypotheses (metaprograms) \( H \) according to the Bayes rule:
\[
P(H \mid D) \propto P(D \mid H) \cdot P(H).
\]

The prior distribution \( P(H) \) integrates two complementary sources of simplicity bias: the meta-program prior \( P_{\mathcal{M}}(H) \) and the induced program prior \( P_{\mathcal{P}}(\widetilde{H}) \). This combined prior is defined as:
\[
P(H) \propto \exp\left(\frac{\ln P_{\mathcal{M}}(H) + \ln P_{\mathcal{P}}(\widetilde{H})}{2}\right),
\]
where \( \widetilde{H} \) denotes the program compiled from the metaprogram \( H \). 

The likelihood \( P(D \mid H) \) measures the consistency of a meta-program \( H \) with the observational data provided, incorporating a noise model to accommodate minor discrepancies between the model predictions and observations.

\paragraph{Formal Language}  
We adopted the Bayesian inference approach from \citep{yang2022} as an optimal reference model to determine the theoretical upper bound on the learnability of formal language rules from the observations generated by our black-boxes or from the interventions queried by LLM. Specifically, we provided strings generated by our formal language black-boxes as observational data to the Bayesian model, which then inferred the underlying symbolic grammar rules.

Just as before, the Bayesian inference framework defines the posterior distribution over candidate hypotheses conditioned on observed data using Bayes' rule:
\[
P(H \mid D) \propto P(D \mid H)\,P(H),
\]
where \( H \) represents a candidate hypothesis (grammar or probabilistic program), \( D \) represents the observed string data generated by the black-box, \( P(H) \) represents the prior probability reflecting initial beliefs about the simplicity and plausibility of hypotheses, and \( P(D \mid H) \) denotes the likelihood of observing data \(D\) given hypothesis \(H\).

The Bayesian model uses a structured prior \( P(H) \), assigning higher probabilities to simpler, more concise grammars or symbolic programs. As observational data increases, Bayesian updating systematically refines prior beliefs into posterior distributions, enhancing the probability assigned to grammars that best explain the data. Formally, each new observed string updates the posterior, shifting probability mass toward hypotheses consistent with the cumulative dataset. By leveraging this Bayesian inference mechanism, we quantify the upper bound of the learnability of the observations, thus providing a rigorous baseline to evaluate LLM's effectiveness in utilizing the same observational data.
\paragraph{Math Equation} 
To infer the parameters of the CES utility model from the observations provided, we followed \citep{foster2019variational} by employing a Bayesian inference approach explicitly conditioned on these observations. Bayesian inference integrates observed data with prior beliefs, updating these beliefs into posterior distributions to progressively improve parameter estimates. Initially, we specified prior distributions for the model parameters:
\begin{align*}
\rho &\sim \text{Beta}(\rho_0, \rho_1), \\
\alpha &\sim \text{Dirichlet}(\alpha_{\text{conc}}), \\
\text{slope} &\sim \text{LogNormal}(\text{slope}_\mu, \text{slope}_\sigma).
\end{align*}

Given pairs of consumption bundles $(d_1, d_2)$ and the corresponding observed user preferences $y$, the Bayesian framework models these preferences probabilistically through a censored sigmoid-normal likelihood:
\[
y \sim \text{CensoredSigmoidNormal}\left(\text{slope}\cdot(U(d_1) - U(d_2)),\;\text{slope}\cdot \text{obs\_sd}\cdot(1+\|d_1 - d_2\|_2)\right),
\]
where $U(d_1)-U(d_2)$ denotes the utility difference between the two bundles. Here, ``censored'' refers to applying a sigmoid function to latent utility values and then truncating the results to the observed preference interval (e.g., $[0, 1]$), ensuring that responses remain within these limits.

The posterior distributions are updated via Bayes' theorem by explicitly integrating observational data:
\[
p(\rho, \alpha, \text{slope}\mid y,d) \propto p(y\mid\rho, \alpha, \text{slope}, d)\, p(\rho, \alpha, \text{slope}),
\]
where $p(\rho, \alpha, \text{slope})$ represents prior distributions and $p(y\mid\rho, \alpha, \text{slope}, d)$ represents the likelihood function given the observations.

While some sources prefer uppercase probability notation such as $P(H \mid D)$, this paper adopts lowercase notation ($p$) consistently for both probability densities and random variables throughout.

Parameter estimation was performed via variational inference \citep{blei2017variational}, iteratively optimizing the evidence lower bound (ELBO), defined as:
\[
\text{ELBO}(\phi) = \mathbb{E}_{q_{\phi}}\left[\log p(y\mid\rho,\alpha,\text{slope},d)\right] - D_{\text{KL}}\left(q_{\phi}(\rho,\alpha,\text{slope})\parallel p(\rho,\alpha,\text{slope})\right),
\]
where $q_{\phi}$ denotes the variational posterior distribution used to approximate the true posterior distribution.

Thus, as additional observational data are incorporated, Bayesian inference continually updates prior beliefs into posterior distributions, systematically refining parameter estimates toward their true underlying values.

\section{Statistical Significant Tests}
\subsection{Repeated-measures ANOVA}
\label{sec:anova}
To statistically evaluate the interaction between models (Bayesian vs. LLM) and steps, we calculated the repeated-measures ANOVAs. Each black-box instance involved multiple repeated measurements corresponding to different steps. Letting \( Y_{ijk} \) represent the performance score for subject \( i \), models \( j \) (Bayesian or LLM), and step \( k \), the repeated-measures ANOVA model can be expressed as:

\[
Y_{ijk} = \mu + S_i + M_j + T_k + (M \times T)_{jk} + \epsilon_{ijk}
\]

where \( \mu \) is the mean in all measurements, \( S_i \) represents the random effect of the subjects (individual seeds), \( M_j \) denotes the main effect of the model, \( T_k \) is the main effect of steps, \( (M \times T)_{jk} \) is the interaction between the model and the step, and \( \epsilon_{ijk} \) represents residual error.

The ANOVA decomposes the total variance into these distinct sources. Specifically, the significance of the interaction of the Step Method $\times$ was determined by calculating the corresponding F-statistic:

\[
F = \frac{MS_{(M \times T)}}{MS_{error}}
\]

where \( MS_{(M \times T)} \) is the mean square for the Method $\times$ Step interaction, and \( MS_{error} \) is the residual mean square. Significance was assessed using an \( F \)-distribution with numerator degrees of freedom equal to \((J-1)(K-1)\), where \( J \) is the number of method levels and \( K \) is the number of steps, and denominator degrees of freedom equal to \((I-1)(K-1)\), where \( I \) is the number of subjects.




\section{Prompts}
\label{sec:prompts}
\subsection{Intervention prompt}
\label{sec:intv_prompt}
\begin{code}
In this task, you are given a ``black box'' and need to determine its inner workings.
{black box information}
You will have a series of turns to interact with the black box. On each turn, you can either gather more information or test your hypothesis. To gather more information, {query instruction}, and obtain a result.
To test your hypothesis, {test instruction}. All the information gathered across all the turns is used to reverse engineer the black box. Throughout the process, you can decide whether the gathered information is sufficient to correctly identify the workings of the black box, in which case you can stop. Otherwise, you need to continue the interaction. Concretely, you can perform one of the following actions at each turn: 1) query, 2) test, or 3) stop.


Provide a *thorough reasoning* before performing the action. Leverage the past observations to design your next query and make your hypothesis as accurate as possible. Below is the format for each action.

Query:
```query
List[int]
```

Test:
```test
List[int]
List[int]
```

Stop:
```stop
```

Note that you should only perform one of the actions above with one input example in your response.

Below are your past observations of the black box:
{observations}
Response:
\end{code}

\subsection{Evaluation Prompts}
\texttt{\underline{\textbf{Program (judge):}}} 
\begin{code}
In this task you will be given a ground truth program and pseudocode that you need to evaluate. You will output a score for the quality of the pseudocode based on a set of assessment criteria.

Below is the ground truth program:
{ground_truth}

Evaluate the quality of the following pseudocode:
{response}

Score the above pseudocode against the ground truth program based on the following criteria (total 10 points):
1. Does the provided pseudocode correctly specify the implementation of the ground truth program and manipulate the variables in the same way? Ignore the programming language difference. [5 point]
2. Does the provided pseudocode specify the implementation in the most simple and straightforward way without extra unused parts (Occam's Razor principle) [5 point]

Explain your judgement and return the final score with the type float and following the format below:
```judgement
YOUR JUDGEMENT HERE
```
```score
YOUR SCORE HERE
```

Response: 
\end{code} 

\texttt{\underline{\textbf{Formal Language (judge):}}}
\begin{code}
In this task, you will be given a ground truth formal language and a proposed rule describing that formal language, which you need to evaluate for quality. You will then output a score based on a set of assessment criteria.

Below is the ground truth formal language:
{ground_truth}

Evaluate the quality of the following formal language rule:
{response}
Score the above formal language rule against the ground truth formal language based on the following criteria (total: 10 points):
1. Does the provided rule correctly generate the examples given in the ground truth? Your score is determined by how many examples are correctly generated out of the total number of examples. [3 points]
2. Does the provided rule correctly reverse engineer the ground truth formal language? [5 point]
3. Is the provided rule in the most simple and straightforward way without extra unused parts (Occam's Razor principle)? Note: If the provided rule is completely incorrect, you should give 0 point for this criterion. [2 point]

Explain your judgement and return the final score with the type float and following the format below:
```judgement
YOUR JUDGEMENT HERE
```
```score
YOUR SCORE HERE
```
  
Response:
\end{code}

\texttt{\underline{\textbf{Math Equation (judge):}}} 
\begin{code}
In this task, you are provided with a ground truth CES utility function and a CES utility function predicted by a model.

Your task is to evaluate the quality of the predicted utility function based on a set of assessment criteria and output a score.

The ground truth utility takes this form:
U(\\mathbf{{x}}) = \\left(\\sum_{{i=1}}^n a_i \\cdot x_i^{{\\text{{rho}}}}\\right)^{{1/\\text{{rho}}}}

The utility depends on the following parameters:
1. a_i: float rounded to the first decimal point and should sum up to 1. (Note that there will be multiple a_i's.)
2. rho: float rounded to the first decimal point.

Below is the information about the ground truth utility function: 
{ground_truth}

Evaluate the quality of the following predicted the parameters of the utility function: 
{response}

Score the predicted utility function against the ground truth using the following criteria (total 10 points):
1. Is the predicted utility function has a correct rho? [2 points]
2. Compare the predicted utility function to the ground truth, how many a_i's are correct (order matters)? This will give us an accuracy percentage. The score for this bullet should be the accuracy percentage times the total allocated 6 points [6 points]
3. In the predicted utility function, do the unknown parameters a_i sum up to 1 and do the number of a_i's match the number of goods? [1 point]
4. Does the predicted utility function express the function in a simple and straightforward way without any unnecessary elements (adhering to the Occam's Razor principle)? [1 point]

Explain your judgement and return the final score with the type float and following the format below:
```judgement
YOUR JUDGEMENT HERE
```

```score
YOUR SCORE HERE
```

Response:
\end{code}

\texttt{\underline{\textbf{Descriptive Evaluation:}}}
\begin{code}
In this task, you are given a ``black box`` and need to determine its inner workings. 
{black box information}

Below are some past observations from the black box:
{observations}

Your task is to reverse engineer the rule underlying {more detailed instructions} in the following format:
```Rule
YOUR RULE HERE
```

Response:
\end{code}

\texttt{\underline{\textbf{Function Implicit Evaluation:}}}
\begin{code}
In this task, you are given a ``black box`` and need to determine its inner workings. 
{black box information}
Below are some past observations from the black box:
{observations}

{More detailed instructions}

Output your generated string in the following format:
```output
YOUR RESPONSE HERE
```

Response:
\end{code}






\section{Reverse Engineering Abilities Across Different Categories of LLMs}
\label{app:diff_llm}
\begin{figure}[h!]
    \centering
    \includegraphics[width=\textwidth]{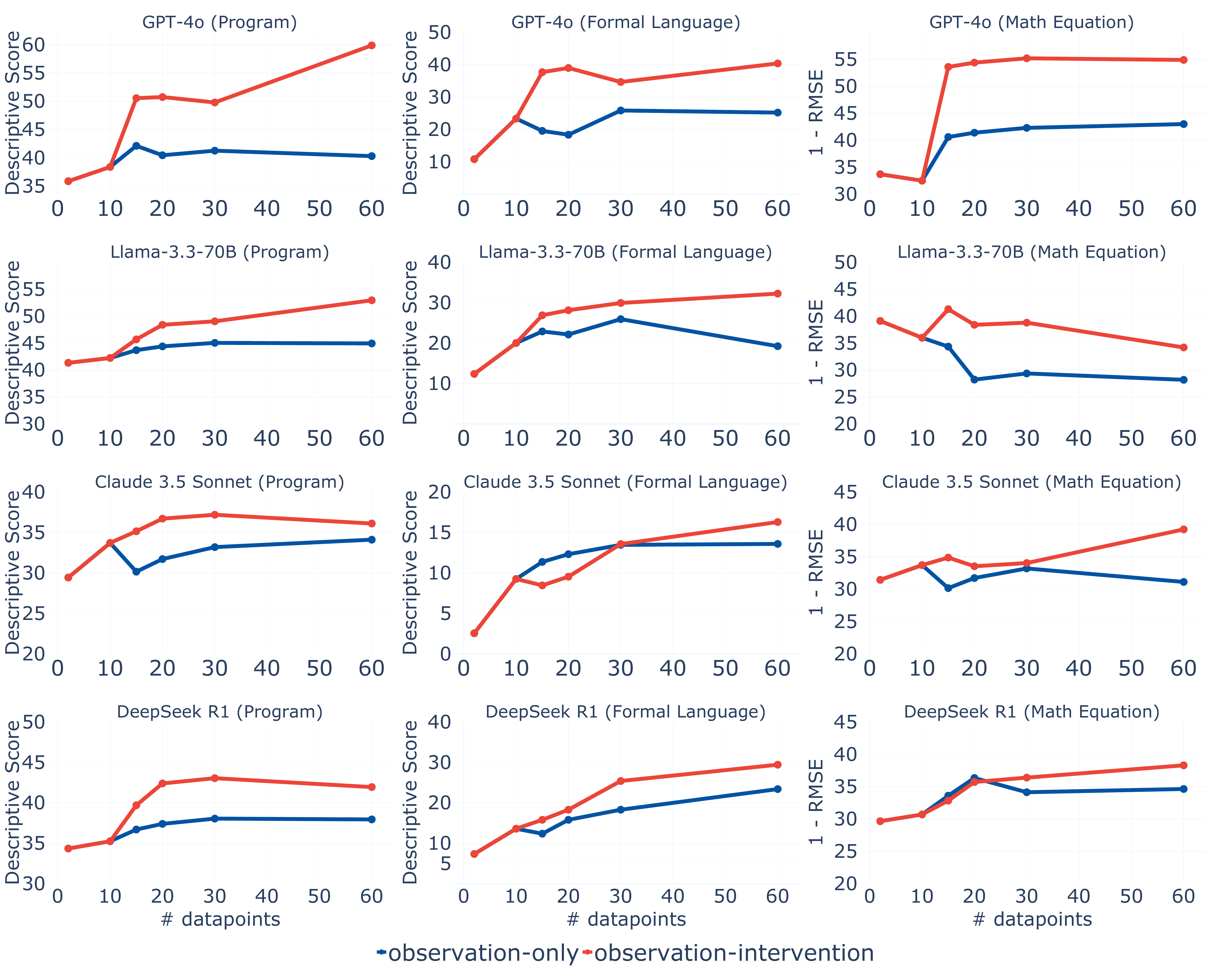}
    \caption{Results of reverse engineering abilities across different categories of LLMs. We report Llama-3.3-70B-Instruct, Claude 3.5 Sonnet and Deepseek R1 on Program, Formal Language and Math Equation.}
    \label{fig:diff_llm}
\end{figure}

In Figure \ref{fig:diff_llm}, we report the results of observation-only and observation + intervention across different LLMs: Llama-3.3-70B-Instruct, Claude-3.5-Sonnet, and DeepSeek-R1. Across nearly all black-box types and models, actively intervening with iteratively refining hypotheses consistently enhances models' understanding of the underlying black-box dynamics. In particular, we show that DeepSeek R1, utilizing Long CoT reasoning, has the potential to continuously extract informative knowledge even from passive learning scenarios. This detailed and long reasoning allows the model to explore various potential hypotheses. However, despite these advantages, DeepSeek R1 does not significantly outperform models without explicit reasoning (e.g., GPT-4o, Llama-3.3-70B-Instruct, Claude 3.5 Sonnet) in reverse-engineering tasks. This finding highlights the inherent limitations of current reasoning steps for existing LLMs.

\section{Common Failure Modes}
\subsection{Human Annotation}
\label{sec:human_expert}
To systematically analyze LLM's failure modes, we defined an LLM reverse-engineering attempt as a failure if its descriptive score was below $2$ out of $10$, according to our descriptive evaluation rubric. For each black-box type, we randomly selected $20$ representative failure cases from the observation-only setting. We have two human experts independently reviewed these examples, categorizing each case based on the nature of the error. Any disagreements were resolved through discussion. Finally, human annotators identified two common failure modes: overcomplication and overlooking.

\subsection{Overcomplication $\&$ Overlooking Examples}
\label{sec:overcomplication_overlooking_examples}

\begin{table}[ht]
\footnotesize
\begin{tabular}{@{}p{2.5cm}p{11.7cm}@{}}
\specialrule{1.2pt}{0pt}{2pt}
\multicolumn{2}{@{}l}{\colorbox{instancepink}{\strut\textbf{Black-box instance: (lambda (singleton (third \$0)))}}} \\
\midrule
\textbf{Observations:} &
Input: \texttt{[97, 53, 5, 33, 65, 62, 51]}; Output: \texttt{[5]} \\
& Input: \texttt{[61, 45, 74, 27, 64]}; Output: \texttt{[74]} \\
& Input: \texttt{[36, 17, 96]}; Output: \texttt{[96]} \\
& Input: \texttt{[79, 32]}; Output: invalid input \\
& Input: \texttt{[90, 77, 18, 39, 12, 93, 9, 87, 42]}; Output: \texttt{[18]} \\
& Input: \texttt{[71, 12, 45, 55, 40, 78, 81, 26]}; Output: \texttt{[45]} \\
& Input: \texttt{[61, 56, 66, 33, 7, 70, 1, 11, 92]}; Output: \texttt{[66]} \\
& Input: \texttt{[90, 100, 85, 80, 0, 78, 63]}; Output: \texttt{[85]} \\
& Input: \texttt{[31, 93, 41, 90, 8, 24]}; Output: \texttt{[41]} \\
& Input: \texttt{[28, 30, 18, 69, 57, 11, 10, 40, 65, 62]}; Output: \texttt{[18]} \\
& Input: \texttt{[38, 70]}; Output: invalid input \\
& Input: \texttt{[90, 15, 70, 42, 69]}; Output: \texttt{[70]} \\
& Input: \texttt{[77, 70, 75, 36]}; Output: \texttt{[75]} \\
& Input: \texttt{[11, 76, 49, 40, 73, 30, 37, 23]}; Output: \texttt{[49]} \\
& Input: \texttt{[23, 4, 78, 84]}; Output: \texttt{[78]} \\
& $\cdots$ \\
& \{More observations\} \\
& $\cdots$ \\
& Input: \texttt{[60, 8, 11, 86, 96]}; Output: \texttt{[11]} \\
& Input: \texttt{[19, 4, 10]}; Output: \texttt{[10]} \\
& Input: \texttt{[87, 50, 90, 67, 35, 66, 30, 27, 86]}; Output: \texttt{[90]} \\
& Input: \texttt{[53, 74, 35, 57, 63, 84, 82, 89, 45, 10]}; Output: \texttt{[35]} \\
& Input: \texttt{[78, 14, 62, 75, 80, 42]}; Output: \texttt{[62]} \\
& Input: \texttt{[31, 2, 93, 34]}; Output: \texttt{[93]} \\
& Input: \texttt{[90, 28]}; Output: invalid input \\
& Input: \texttt{[21, 42, 54, 7, 12, 100]}; Output: \texttt{[54]} \\
& Input: \texttt{[89, 28, 5]}; Output: \texttt{[5]} \\
& Input: \texttt{[81, 68, 77, 87, 9, 3, 15, 81, 24, 77]}; Output: \texttt{[77]} \\
& Input: \texttt{[15, 50, 11, 47, 14, 4, 77, 2, 24, 23]}; Output: \texttt{[11]} \\
& Input: \texttt{[61, 26]}; Output: invalid input \\
& Input: \texttt{[86]}; Output: invalid input \\
& Input: \texttt{[79, 12, 33, 8, 28, 9, 82]}; Output: \texttt{[33]} \\
& Input: \texttt{[44, 55, 23, 7, 64]}; Output: \texttt{[23]} \\
\addlinespace[0.5em]
\midrule

\begin{minipage}[t]{\linewidth}
\textbf{Model response:}\\
\textbf{(Overcomplication)}
\end{minipage}

&
\begin{tcolorbox}[colframe=black, colback=modelbox, sharp corners, boxrule=0.7pt, boxsep=3pt]
FUNCTION black\_box(input\_list):\\
\ \ \ IF length of input\_list < 3:\\
\ \ \ \ \ \ RETURN "invalid input"\\
\ \ \ ELSE:\\
\ \ \ \ \ \ \textbf{\textcolor{myred}{SORT input\_list in ascending order}}\\
\ \ \ \ \ \ RETURN the third element in the sorted input\_list
\end{tcolorbox}

\begin{tcolorbox}[colframe=black, colback=modelbox, sharp corners, boxrule=0.7pt, boxsep=3pt]
Explanation:

The black box function takes a list of integers as input. If the list contains fewer than 3 elements, it returns "invalid input". Otherwise, \textbf{\textcolor{myred}{it sorts the list in ascending order and returns the third smallest element}} (which is the third element in the sorted list).
\end{tcolorbox}
\\
\specialrule{0.2pt}{0pt}{0pt}
\end{tabular}
\caption{Overcomplication example of Program. Red: LLM comes up with a more complex rule of black-box instance.}
\label{table:prog_overcomp}
\end{table}

\begin{table}[ht]
\footnotesize
\begin{tabular}{@{}p{2.5cm}p{11.7cm}@{}}
\specialrule{1.2pt}{0pt}{2pt}
\multicolumn{2}{@{}l}{\colorbox{instancepink}{\strut\textbf{Black-box instance: AnBmAnBm ($A^nB^mA^nB^m$)}}} \\
\midrule
\textbf{Observations:} &
AAAAAAAAAAAAAA is generated by the black box \\
& AABBBBBBAABBBBBB is generated by the black box \\
& AAAAAAAAABBAAAAAAAAABB is generated by the black box \\
& AAAAAAAABBBBBAAAAAAAABBBBB is generated by the black box \\
& AAAAABBBBBBBBAAAAABBBBBBBB is generated by the black box \\
& AAAAAAAABBBBBBBBBBAAAAAAAABBBBBBBBBB is generated by the black box \\
& AAABBBBBBBBBAAABBBBBBBBB is generated by the black box \\
& AAAAAAAABBBBBBAAAAAAAABBBBBB is generated by the black box \\
& AAABBBBBBBBBAAABBBBBBBBB is generated by the black box \\
& $...$ \\
& \{More observations\} \\
& $...$ \\
& AAABBBBAAABBBB is generated by the black box \\
& AAAABBAAAABB is generated by the black box \\
& AAAAAAAAABAAAAAAAAAB is generated by the black box \\
& AAAAAABBBBBBAAAAAABBBBBB is generated by the black box \\
& AAAAAABBBBBAAAAAABBBBB is generated by the black box \\
& AAAAAAAAAAAAAAAAAA is generated by the black box \\
& AAAAAAAAAABBBBBBBBAAAAAAAAAABBBBBBBB is generated by the black box \\
& AAAAAAAAAABBAAAAAAAAABB is generated by the black box \\
& AAAAABBBBBBBBBAAAAABBBBBBBBB is generated by the black box \\
& AAAAABBBBBBBBAAAAABBBBBBBB is generated by the black box \\
& AAAABBBBAAAABBBB is generated by the black box \\
\addlinespace[0.5em]
\midrule

\begin{minipage}[t]{\linewidth}
\textbf{Model response:}\\
\textbf{(Overcomplication)}
\end{minipage}

&
The formal language generates strings composed of two segments: a sequence of 'A's followed by a sequence of 'B's.
\textbf{\textcolor{Maroon}{The number of 'A's in the first segment must be equal to the number of 'B's in the second segment,}}
or the number of 'A's can be zero, resulting in a string of only 'B's. Additionally, the language allows for strings composed entirely of 'A's or entirely of 'B's. The sequences of 'A's and 'B's can be repeated multiple times, as long as the rule of equal counts between 'A's and 'B's in each segment is maintained. \\
\specialrule{1.2pt}{2pt}{0pt}
\end{tabular}
\caption{Overcomplication example for Formal Language. Red: The LLM predicts a rule that is more complex than necessary for this black-box instance. In this example, it incorrectly reverse the number of As to be equal to the number of Bs.}
\label{table:lang_overcomp}
\end{table}
\begin{table}[ht]
\footnotesize
\begin{tabular}{@{}p{2.5cm}p{11.7cm}@{}}
\specialrule{1.2pt}{0pt}{2pt}
\multicolumn{2}{@{}l}{\colorbox{SkyBlue!30}{\strut\textbf{Black-box instance: XXX (three copies of the same string)}}} \\
\midrule
\textbf{Observations:} 
&BBABBABABBABBABABBABBABA is generated by the black box\\
&BABBABBABBABBABBAB is generated by the black box\\
&ABBABBABB is generated by the black box\\
&ABAAAABABAAAABABAAAAB is generated by the black box\\
&AAA is generated by the black box\\
&AAAAAA is generated by the black box\\
&BBB is generated by the black box\\
&ABAABAABAABAABAABA is generated by the black box\\
&ABABAB is generated by the black box\\
&ABAABAABA is generated by the black box\\
& $...$ \\
& \{More observations\} \\
& $...$ \\
&BBBBAABABABBBBAABABABBBBAABABA is generated by the black box\\
&ABABAB is generated by the black box\\
&AAABAABAAABAABAAABAAB is generated by the black box\\
&ABABAB is generated by the black box\\
&BABBBBBBBABABBBBBBBABABBBBBBBA is generated by the black box\\
&ABBBABBBABBB is generated by the black box\\
&BABABA is generated by the black box\\
&BBABAABBABBABAABBABBABAABBA is generated by the black box\\
&AAAAAA is generated by the black box\\
\addlinespace[0.5em]
\midrule

\begin{minipage}[t]{\linewidth}
\textbf{Model response:}\\
\textbf{(Overlooking)}
\end{minipage}

&
\begin{minipage}[t]{\linewidth}
The black box generates strings that are composed of repeated patterns or segments. Each string can be broken down into one or more segments, where each segment is a repetition of a smaller pattern. The patterns can be composed of the letters 'A' and 'B', and the repetition of these patterns can vary in length. Additionally, the black box can generate an empty string.
\\ \\
The key characteristics of the generated strings are:\\
1. The strings can be empty. \\
2. The strings can be composed of repeated patterns of 'A' and 'B'.\\
3. The patterns themselves can vary in complexity and length.\\
4. The repetition of these patterns can be consistent or vary within the string.
\\ \\
Examples of patterns include: \\
- Single letters repeated (e.g., "AAA", "BBB"). \\
- Simple alternating patterns (e.g., "ABABAB", "BABABA"). \\
- More complex patterns with varying repetition (e.g., "ABBABBABB", "BABBABBABBABBABBAB").\\ \\
\textbf{\textcolor{MidnightBlue}{
The rule can be estimated as: The black box generates strings by repeating patterns of 'A' and 'B', where the patterns can be simple or complex, and the repetition can be consistent or vary within the string.}}
\end{minipage}
\\
\specialrule{1.2pt}{2pt}{0pt}
\end{tabular}
\caption{Overlooking example of Formal Language. LLM estimates a rule of the black-box instance without utilizing the observed data to come up with the hypothesis.}
\label{table:lang_overlook}
\end{table}
\begin{table}[h!]
\footnotesize
\begin{tabular}{@{}p{2.5cm}p{11.7cm}@{}}
\specialrule{1.2pt}{0pt}{2pt}
\multicolumn{2}{@{}l}{\colorbox{SkyBlue!30}{\strut\textbf{Black-box instance: alpha=[0.1,0.1,0.4,0.4]; rho=0.4}}} \\
\midrule
\textbf{Observations:} 
&\texttt{Basket1: [87.79948556831374, 39.184851241810016, 22.551465649464433, 42.84633678341351];} \\
&\texttt{Basket2: [66.20150946367784, 80.43281524767865, 68.02177295663886, 8.112415033884117];} \\
&\texttt{Preference: Basket1} \\
&\texttt{Basket1: [20.086807088533046, 79.79230548488334, 8.817190035426963, 90.22168650956948];} \\
&\texttt{Basket2: [90.40796323190197, 70.51080246059688, 69.96510352125082, 48.476064999077785];} \\
&\texttt{Preference: Basket2} \\
&\texttt{Basket1: [8.003392409139845, 35.23615132704936, 76.3120116750855, 59.18156984237496];} \\
&\texttt{Basket2: [31.278783818535036, 18.047644469275518, 21.855445800494433, 53.88293391636068];} \\
&\texttt{Preference: Basket1} \\
& $...$ \\
& \{More observations\} \\
& $...$ \\
&\texttt{Basket1: [40.80654893485903, 14.795240036848078, 37.35074210377093, 79.44298024697972];} \\
&\texttt{Basket2: [51.30239830060435, 50.0604427932334, 21.59530896237239, 36.88204913376821];} \\
&\texttt{Preference: Basket1} \\
&\texttt{Basket1: [75.18968775093136, 42.30088453163497, 18.388385810882728, 95.92185012513752];} \\
&\texttt{Basket2: [93.37477559149177, 86.19812315165629, 49.438229718409715, 98.68538582332832];} \\
&\texttt{Preference: Basket2} \\
&\texttt{Basket1: [70.30309661179918, 94.48475506394722, 78.28267517818938, 70.90519145824096];} \\
&\texttt{Basket2: [63.648862012782175, 99.80230448540196, 61.33979160499423, 78.38413565179029];} \\
&\texttt{Preference: Basket1} \\
\addlinespace[0.5em]
\midrule

\begin{minipage}[t]{\linewidth}
\textbf{Model response:}\\
\textbf{(Overlooking)}
\end{minipage}

&
\begin{minipage}[t]{\linewidth}
To reverse engineer the CES utility function, we need to determine the parameters \( a_i \) and \( \rho \) that best fit the observed preferences between baskets. The CES utility function is given by:
\[
U(\mathbf{x}) = \left(\sum_{i=1}^n a_i \cdot x_i^{\rho}\right)^{1/\rho}
\]
where \( a_i \) are the weights for each good and \( \rho \) is the elasticity parameter.

Steps to Determine Parameters:
\begin{itemize}
\item Analyze Preferences: For each pair of baskets, the basket with the higher utility is preferred.
\item Estimate \( \rho \): The parameter \( \rho \) affects the degree of substitutability between goods. \( \rho = 1 \) is perfect substitutes, \( \rho = 0 \) is Cobb-Douglas, \( \rho < 0 \) is complementarity.
\item Estimate \( a_i \): The weights \( a_i \) sum to 1 and reflect the importance of each good.
\item Iterative Fitting: Use optimization to adjust \( a_i \) and \( \rho \) to minimize preference violations.
\end{itemize}

\textbf{\textcolor{MidnightBlue}{Estimation: After analyzing the data and testing various combinations, the following provides a good fit for the observed preferences:}}

Best Fit Utility Function:
\[
U(\mathbf{x}) = \left(0.3 x_1^{0.5} + 0.2 x_2^{0.5} + 0.3 x_3^{0.5} + 0.2 x_4^{0.5}\right)^{2}
\]
This is derived by fitting the parameters to minimize preference violations. The weights \( a_i \) reflect the relative importance of each good, and \( \rho = 0.5 \) is selected to match substitutability.

\end{minipage}
\\
\specialrule{1.2pt}{2pt}{0pt}
\end{tabular}
\caption{Overlooking example of Math Equation. Instead of explicit plug-in all observed data into the utility function, LLM estimates $\alpha$ and $\rho$.}
\label{table:ces_overlook}
\end{table}

Across the three black-box types, we find that overcomplication is a common failure mode, particularly in the Program, while overlooking most often occurs in Math Equation. For Formal Language, both overcomplication and overlooking are observed when LLMs fail at reverse engineering. In Tables \ref{table:prog_overcomp},\ref{table:lang_overcomp} \ref{table:lang_overlook} and \ref{table:ces_overlook}, we show the failure examples for Program (overcomplication), Formal Language (overcomplication \& overlooking) and Math Equation (overlooking).

\section{Complexity Categorization}
\label{app:complexity_cat}
We rank the complexity level from $1-5$. Each black-box type includes multiple instances of varying task complexity.
\paragraph{Program. }The complexity level is determined based on the number of operations, which ranges from $1-12$. Instances with fewer than $2$ operations are classified as complexity level $1$ ($complexity-1$), those with fewer than $4$ operations as $complexity-2$, fewer than $6$ operations as $complexity-3$, and fewer than $8$ operations as $complexity-4$. Due to the limited number of remaining examples, all others are grouped into the highest complexity level ($complexity-5$).
\paragraph{Formal Language.} Instead of using five complexity levels, we divided the Formal Language instances into three levels, drawing on insights from \citep{la2007robust}. Specifically, we categorized regular language instances as complexity-1 black-boxes, context-free languages as complexity-3, and context-sensitive languages as complexity-5.

\paragraph{Math Equation.} We categorize complexity levels according to the number of goods involved, ranging from $2$ to $6$. Specifically, instances with $2$ goods are labeled as $complexity-1$, $3$ goods as $complexity-2$, and so on, with instances involving 6 goods classified as the highest complexity level, $complexity-5$.

\section{Evaluation of the Reverse-Engineering Ability}
\label{sec:func_eval}

\begin{figure}[h!]
    \centering
    \includegraphics[width=0.95\textwidth]{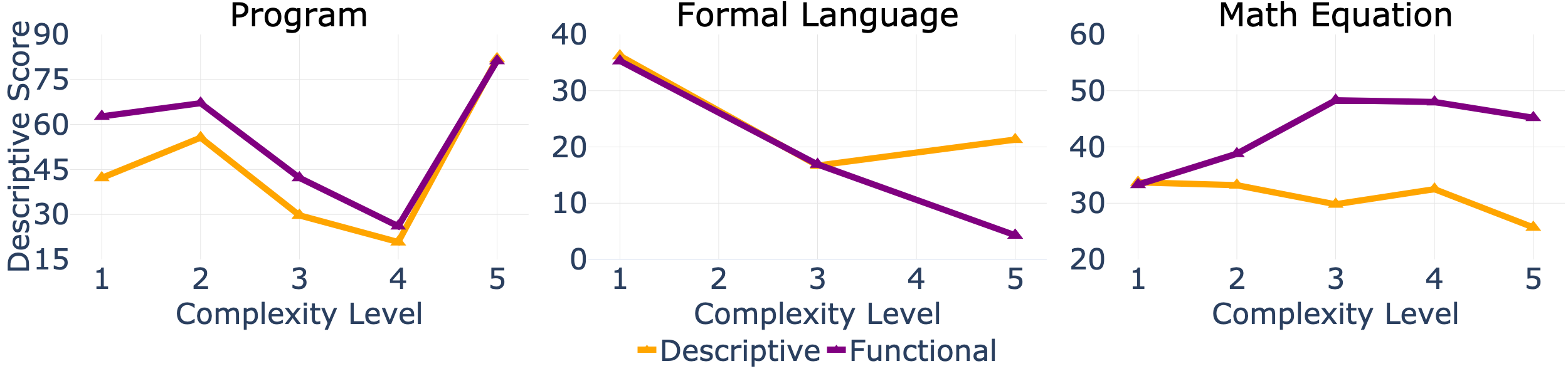}
    \caption{Comparison of descriptive evaluation (yellow) and functional evaluation (purple) across black-box complexity levels.}
    \label{fig:desc_vs_func_eval}
\end{figure}

Unlike typical tasks used to benchmark LLMs, such as solving math problems or question answering which are commonly evaluated using accuracies, the reverse-engineering ability is less straightforward.
One can assess how well the black-box $f^*$ is recovered by an LLM by: 1) \textit{descriptive} evaluation where the LLM verbalizes the hypothesis to compare to the ground truth and 2) \textit{functional} evaluation which captures how well the LLM emulates the black-box's input-output dynamics and generalizes to unseen examples \citep{kang2024learning}. 
In functional evaluation, the LLM directly predicts the response conditioned on the test query and the past observations and compute accuracy $\texttt{Acc} = \frac{1}{M} \sum_{i=1}^M \mathbbm{1}[ y_i^\text{test} = \mathcal{M}(x_i^{\text{test}}, \mathcal{O}) ]$, without generating the black-box implementation, akin to in-context learning \citep{brown2020language}.
As shown in Figure~\ref{fig:desc_vs_func_eval}, descriptive and functional evaluation trends align for Program across complexity levels. However, we also observe discrepancies of trends between the two evaluations for Formal Language (complexity level 3 to 5) and Math Equation (complexity level 1 to 3), demonstrating that the evaluation protocol and the \textit{format} of the model output may capture different strengths and weaknesses of the model.
For Program, we used the original samples from the black box of the list mapping program as test cases \citep{rule2024symbolic} and ensured that none of these input–output pairs were included in the observations. For Formal Language and Math Equation, we use our deterministic black-box randomly sample 20 test cases per black-box instance.


\section{Another Black-Box Type: Board Game}
\subsection{Black-Box Design}
We design a connect-$n$ board game ($2 \times 2$ to $9 \times 9$) variant following \citep{zhang2024people}. The black-box is defined by the rules that dictate the winning condition of the game (e.g., ``Win by connecting 3 stones in a column.''). The LLM can query with a board state and an action, and the black-box responds with the new board state and a game status (win/lose/draw/ongoing). In our black-box design, every game instance exposes two modes—\texttt{observation} (observation-only) and \texttt{intervention} (observation-intervention) —and uses the
symbols \texttt{X} and \texttt{O} to mark the moves of the two players.

\paragraph{Game definition.}
For a given instance, let the board be a \(r\times c\) grid and let
\(\langle r_\text{win},\,c_\text{win},\,d_\text{win}\rangle\) denote the required number of consecutive
marks needed to win horizontally, vertically, and diagonally, respectively.
During play the black-box tracks the current board state \(B\),
the active player \(p\in\{\texttt{X},\texttt{O}\}\), and whether the game has ended.

In \texttt{observation} mode, an external LLM supplies an \emph{initial} board (or leaves it empty).
The black-box generates the following as the outputs:
\begin{itemize}\setlength\itemsep{2pt}
  \item the round number,
  \item the updated board,
  \item whose move it was last,
  \item the current game status (\textit{ongoing}, \textit{draw}, \textit{PlayerX\_wins}, etc.).
\end{itemize}
If the move ends the game, the record also names the winner.

In \texttt{intervention}, the LLM needs to specify
(i) additional pieces to place on the board,
(ii) the candidate action it wishes the black-box to take, and
(iii) optionally, a predicted follow-up board.
The black-box returns the same structured record as in observation mode.
If the LLM also proposed a prediction of the next state, the black-box confirms it (\textit{``Correct''}) or explains why it is invalid. For Board Game, we do not have a Bayesian model as the optimal reference for the comparison.


\subsection{GPT-4o Results}
\begin{figure}[h!]
    \centering
    \includegraphics[width=0.6\textwidth]{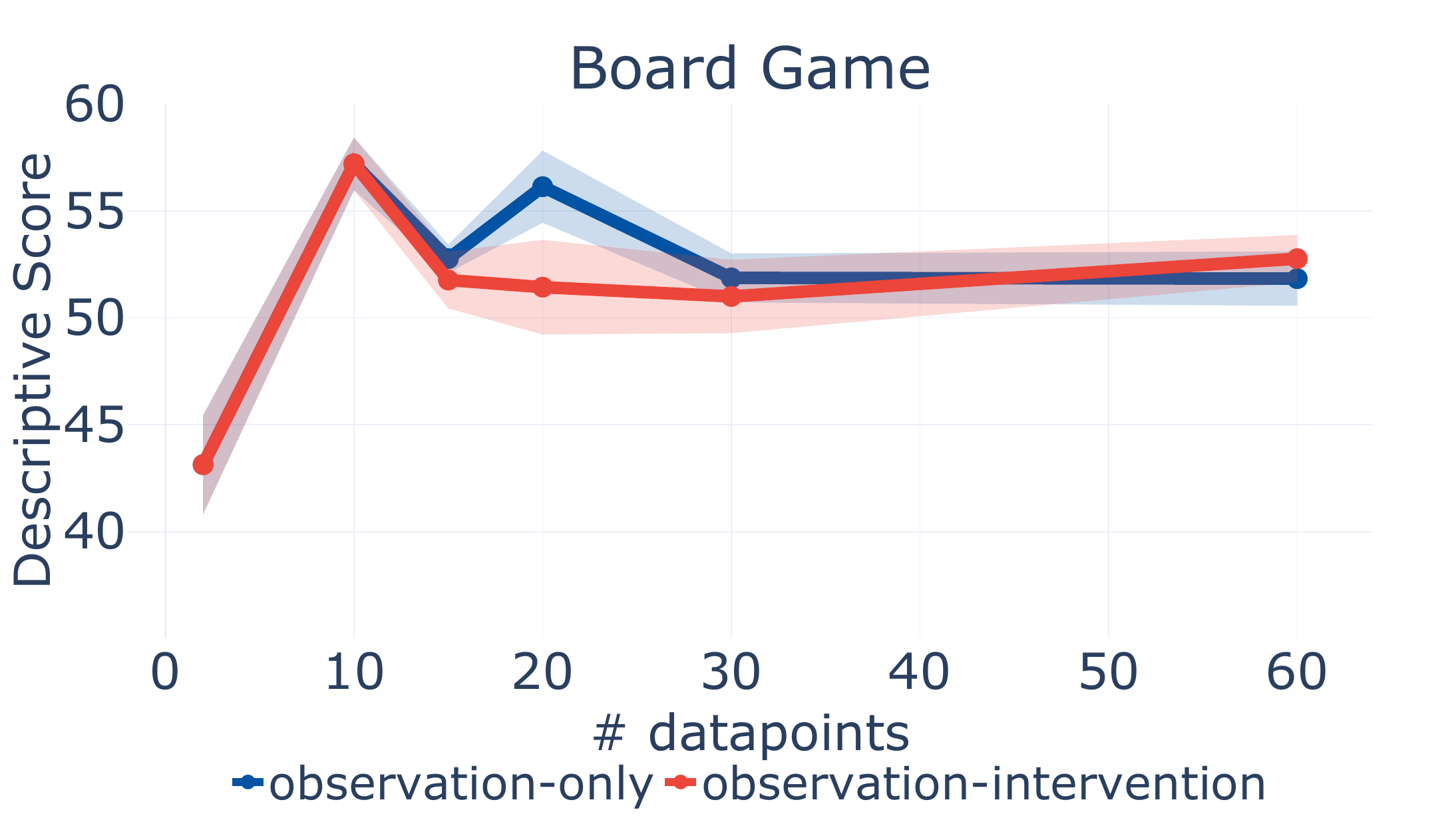}
    \caption{Observation-only and observation-intervention results for Board Game.
    }
    \label{fig:obs_intv_board_game}
\end{figure}

In Figure~\ref{fig:obs_intv_board_game}, we do not observe the same trends seen in Programs, Formal Language, and Math Equation black-box types. For Board Game, actively collected data does not improve the reverse-engineering performance of the model, indicating that the data gathered is not even significantly informative for the LLM itself. We hypothesize that this is because, to query the black-box, the LLM must (1) generate a board state, (2) propose a next move, and (3) predict the resulting board state, requiring a multi-step reasoning process. These compounded requirements make it challenging for the LLM to probe edge cases or effectively reduce uncertainty about the black-box. This result further highlights a key limitation of current LLMs: When the information signal from the black-box is sparse, actively collected data remain of limited utility.


\section{Functional Evaluation Details}\label{app:func_eval}
For Program, we used the original samples from the black box of the list mapping program as test cases \citep{rule2024symbolic} and ensured that none of these input–output pairs were included in the observations. For Formal Language and Math Equation, we use our deterministic black-box randomly sample 20 test cases per black-box instance.

\section{Complexity Categorization}
\label{app:complexity_cat}
We rank the complexity level from $1-5$. Each black-box type includes multiple instances of varying task complexity.
\paragraph{Program. }The complexity level is determined based on the number of operations, which ranges from $1-12$. Instances with fewer than $2$ operations are classified as complexity level $1$ ($complexity-1$), those with fewer than $4$ operations as $complexity-2$, fewer than $6$ operations as $complexity-3$, and fewer than $8$ operations as $complexity-4$. Due to the limited number of remaining examples, all others are grouped into the highest complexity level ($complexity-5$).
\paragraph{Formal Language.} Instead of using five complexity levels, we divided the Formal Language instances into three levels, drawing on insights from \citep{la2007robust}. Specifically, we categorized regular language instances as complexity-1 black-boxes, context-free languages as complexity-3, and context-sensitive languages as complexity-5.

\paragraph{Math Equation.} We categorize complexity levels according to the number of goods involved, ranging from $2$ to $6$. Specifically, instances with $2$ goods are labeled as $complexity-1$, $3$ goods as $complexity-2$, and so on, with instances involving 6 goods classified as the highest complexity level, $complexity-5$.

\end{document}